\documentclass[10pt,twocolumn,letterpaper]{article}

\usepackage[pagenumbers]{cvpr} %

\usepackage{pifont}
\newcommand{\cmark}{\textcolor{green}{\ding{51}}}
\newcommand{\xmark}{\textcolor{red}{\ding{55}}}

\usepackage{color, colortbl}
\definecolor{Gray}{gray}{0.93}
\definecolor{Orange}{rgb}{1,0.5,0}
\definecolor{DGray}{gray}{0.83}
\definecolor{LightCyan}{rgb}{0.88,1,1}

\definecolor{WarnREd}{rgb}{1,0.4,0.4}
\definecolor{WarnOrange}{rgb}{1,0.682,0.502}
\definecolor{WarnPink}{rgb}{0.9176, 0.7215, 0.7215}
\definecolor{GoodGreen}{rgb}{0.5019, 0.9215, 0.6039}

\usepackage[T1]{fontenc}

\definecolor{styleblue}{HTML}{504099}
\definecolor{mypurple}{HTML}{9391ff}

\usepackage{algpseudocode}
\algrenewcommand\algorithmicrequire{\textbf{Input:}}
\algrenewcommand\algorithmicensure{\textbf{Return:}}

\usepackage{nth}

\usepackage{wrapfig}

\usepackage{mathtools}

\usepackage{pifont}
\usepackage{color, colortbl}

\usepackage{blindtext}
\usepackage{lipsum}
\usepackage{multirow}
\usepackage{listings}

\usepackage{bbm}

\usepackage [english]{babel}
\usepackage [autostyle, english = american]{csquotes}

\usepackage{pifont}
\usepackage{url}
\usepackage[most]{tcolorbox}

\usepackage{lipsum}
\usepackage{soul}
\usepackage{xcolor}
\usepackage{wrapfig}
\usepackage{multirow,mathtools } 

\usepackage{adjustbox}
\MakeOuterQuote{"}

\usepackage{microtype}
\usepackage{graphicx}
\usepackage{booktabs} %

\usepackage{amsmath}

\usepackage{tablefootnote}

\usepackage{threeparttable}
\usepackage[normalem]{ulem}

\usepackage{amsmath,amsfonts,bm}

\def\eqref#1{(\ref{#1})}

\def\1{\bm{1}}

\DeclareMathAlphabet{\mathsfit}{\encodingdefault}{\sfdefault}{m}{sl}
\SetMathAlphabet{\mathsfit}{bold}{\encodingdefault}{\sfdefault}{bx}{n}

\newcommand{\ours}{\textsc{StreamForce}}

\usepackage{stfloats}
\usepackage{placeins}
\usepackage[ruled]{algorithm2e}

\SetAlFnt{\small}
\SetAlCapFnt{\small}
\SetAlCapNameFnt{\small}
\SetAlCapHSkip{0pt}

\providecommand{\Description}[1]{}

\definecolor{cvprblue}{rgb}{0.21,0.49,0.74}
\usepackage[pagebackref,breaklinks,colorlinks,allcolors=cvprblue]{hyperref}

\def\paperID{*****} %
\def\confName{CVPR}
\def\confYear{2026}

\title{Streaming Video Generation with Streaming Force Control}

\author{
    Hanhui Wang$^{1*}$, 
    Yiming Xie$^{1,2*}$, 
    Haiwen Feng$^{2,3}$, 
    Zhaoyang Lv$^{2}$, \\
    Shenlong Wang$^{4}$$^{\dagger}$, 
    Huaizu Jiang$^{1}$$^{\dagger}$\\[0.2cm]
    $^1$Northeastern University \quad
    $^2$Impossible Research \quad
    $^3$University of California, Berkeley \\
    $^4$University of Illinois Urbana-Champaign \\[0.2cm]
    {\small $^*$ Equal contribution\quad $^{\dagger}$ Equal advising}
}

\begin{document}
\twocolumn[{%
\renewcommand\twocolumn[1][]{#1}%

\maketitle

\begin{center}
    \vspace*{-1.em}
    \includegraphics[width=0.94\textwidth]{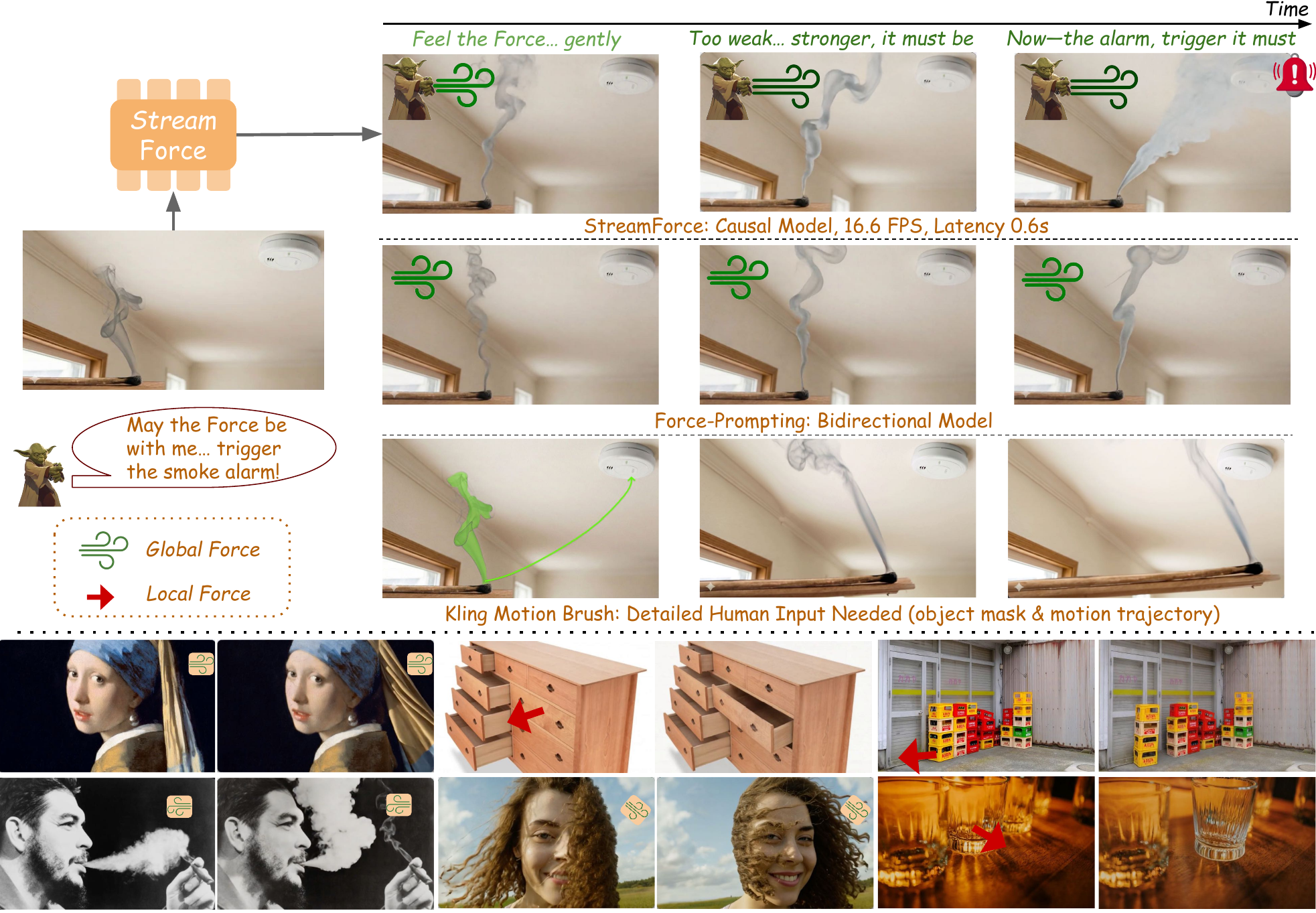}
    \vspace*{-1.em}
    \captionof{figure}{{\bf \ours} enables users to generate streaming force-conditioned videos from a single image by predicting future scene dynamics. Motion is generated online as streaming forces are applied, and users can modify forces at any time to steer the evolving video. 
        We show the \includegraphics[height=1em]{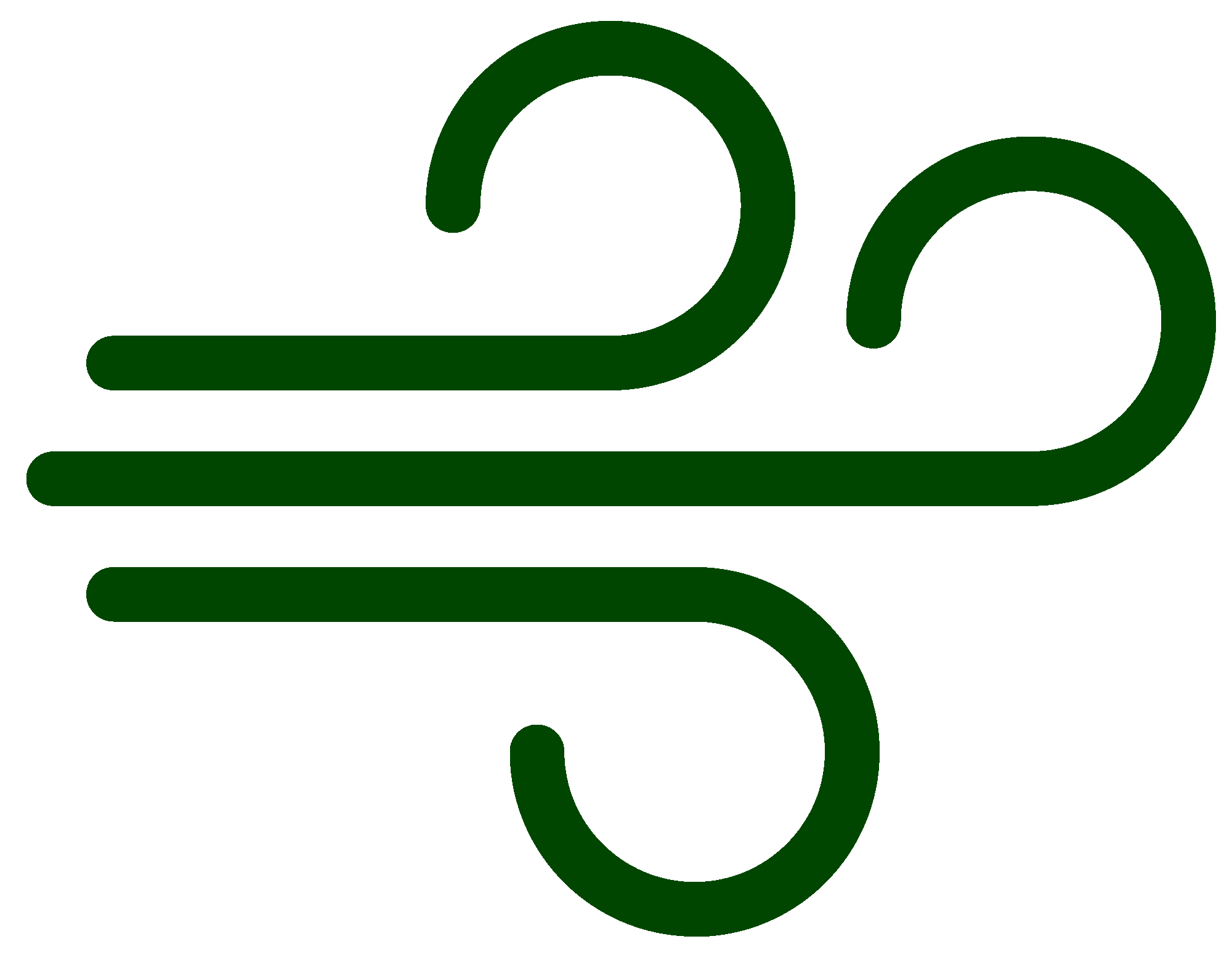} Global Force (affecting the entire scene like wind) and \includegraphics[height=1em]{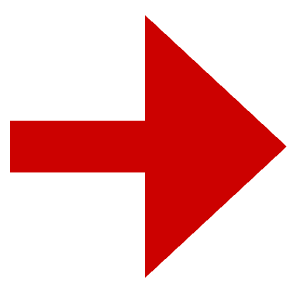} Local Force (acting on specific regions like a push) examples in the bottom part.}
    \label{fig:teaser}
\end{center}
}]

\begin{abstract}
We introduce \ours, a streaming video generation framework that enables physically grounded control through continuous force inputs.
Unlike prior video models that train separate models for different force types, assume fixed forces, or rely on non-causal processing, \ours~is a causal and unified model that responds instantly and coherently to both local and global, time-varying forces. 
To achieve this, we design a unified force representation as a control signal and develop a distillation pipeline for force-controllable video generation. Our model combines autoregressive efficiency with force responsiveness, sustaining stable photometric and dynamic realism.
\ours~runs at up to 16.6 FPS on a single GPU, achieving state-of-the-art performance in both force adherence and motion realism. Project website: \href{https://neu-vi.github.io/StreamForce/}{https://neu-vi.github.io/StreamForce/}
\end{abstract}

\section{Introduction}
\label{sec:intro}
Images capture moments, and videos replay fixed dynamics. Yet the world contains infinitely many possibilities.  Recent progress in diffusion-based video generation~\cite{ho2022imagen, singer2022make, blattmann2023align, blattmann2023stable, hong2022cogvideo, peebles2023scalable, gupta2024photorealistic, openai2024sora, polyak2024movie, yang2024cogvideox, kong2024hunyuanvideo, wan2025wan} has begun to push these models toward a new role: acting as \textit{interactive world models}~\cite{ball2025genie, bar2025navigation, he2025matrix, li2025hunyuan, parkerholder2024genie2, po2025long, ye2025yan, zhang2025matrix}. 
Given an input image, one may wish to ask: \emph{what would happen if we apply an action to the scene?} 
Imagine sliding a ball across a table to a marked spot: you push it, see it drifting left, and apply a corrective nudge to guide it back on course.
To enable such capabilities, two key desiderata arise:  \textbf{force interaction}, where a model can respond to natural and physically meaningful force control inputs; and 
\textbf{streaming generation}, where video unfolds sequentially, allowing users to observe results and adjust forces at any point during generation.

Existing approaches fail to satisfy both requirements simultaneously. 
Various control signals have been explored for diffusion-based video generation.
Text control~\cite{wan2025wan,yang2024cogvideox,blattmann2023stable,hacohenLTX2EfficientJoint2026, openai2024sora,kong2024hunyuanvideo} is intuitive but coarse; camera or actor control~\cite{he2024cameractrl,bai2025recammaster,yu2025trajectorycrafter,yang2024direct,wangMotionCtrlUnifiedFlexible2023, xie2024sv4d} enables viewpoint manipulation but lacks fine-grained interaction; motion control~\cite{shin2025motionstream,nam2025optical} is expressive but difficult for users to specify naturally, e.g., for a fluid dynamics as shown in \textbf{\cref{fig:teaser}} (Kling Motion Brush results). 
More fundamentally, trajectory-based control specifies the effect (where objects move), requiring users to predetermine the motion. 
It cannot express that the same action should produce different outcomes depending on the object: a push moves a heavy object slowly and a light one quickly. 
Force-based control instead specifies the cause, delegating the physical response to the model, so that object-dependent dynamics emerge
naturally without explicit material or mass specification. 
Recent causal
autoregressive models~\cite{shin2025motionstream, yin2025slow, huang2025self, po2025long,
liu2025rolling}, which distill from bidirectional teacher models, have brought trajectory-based
control to real-time streaming speeds, but remain limited to kinematic
inputs, lacking mechanisms for modeling physically plausible interactions such as force inputs.

Force-Prompting~\cite{gillman2025force} is the first work to explore
force-conditioned video generation, but several limitations remain.
First, it treats global forces (such as wind acting on an entire scene) and local forces (such as pushing a specific object) as fundamentally different problems, using separate representations and training separate models for each. 
This not only doubles the modeling cost but also precludes learning shared dynamics across force types. 
Second, its training data contains only fixed forces that remain constant throughout each video, providing no supervision for temporally varying inputs. 
As a result, the model cannot respond coherently when a user changes force direction or magnitude during generation.
Third, it relies on bidirectional diffusion that generates the entire sequence offline, requiring all forces to be specified upfront before generation begins. This prevents users from adjusting forces based on observed motion, eliminating the interactive feedback loop that makes force control compelling, as shown in \textbf{\cref{fig:teaser}} (Force-Prompting results).

In this paper, we present \textbf{\ours}, a streaming video generation framework that enables force-driven interaction through continuous force inputs. 
We introduce a unified force representation that encodes both local and global forces within a single formulation. 
Rather than treating different force types as separate problems, our representation uses a shared pixel-aligned masked force map, enabling a single model to handle diverse physical interactions and learn shared dynamics across force types. 
We further construct a force-conditioned dataset with temporally varying forces, exposing the model to dynamic force transitions during training.
More importantly, to enable streaming generation, we distill a force-conditioned bidirectional teacher into a causal autoregressive student. 
We find that naively introducing force conditioning into existing distillation pipelines leads to unrealistic dynamics and weak responsiveness, as standard distillation does not preserve the force-motion correspondence learned by the teacher (see~\cref{experiments}). 
To overcome this, we develop a force-aware distillation paradigm that enforces force conditioning throughout the entire process, using diverse image-force data to preserve both controllability and open-domain visual generalization. 
Together, these enable \ours\ to generate videos causally while responding coherently to spatially and temporally varying forces.

\ours~runs at 16.6 FPS at a resolution of 832$\times$480 with 0.6-second latency on a single H200 GPU. It achieves state-of-the-art performance in both force adherence and motion realism. These results enable online, physically plausible interaction with generated videos, where users can apply forces and immediately observe the resulting motion. More broadly, \ours~moves generative video models closer to interactive world models capable of predicting and manipulating dynamic physical environments.

Our contributions are threefold: 
(1) We introduce \textbf{\ours}, an autoregressive video generation framework that enables interaction through continuous force inputs. 
(2) We propose a \textbf{unified force representation} that models both local contact forces and global environmental forces within a single formulation. 
(3) We develop a \textbf{force-aware distillation pipeline and dataset} that enable autoregressive video models to learn and respond to dynamic force changes.

\section{Related Work}
\label{sec:related_work}

\noindent\textbf{Bidirectional Video Generation.}
Most state-of-the-art video generators today are built upon bidirectional diffusion models, which jointly denoise all frames conditioned on both past and future temporal context. Recent architectures have evolved from early space-time U-Nets~\cite{blattmann2023stable, hong2022cogvideo} to DiT-style transformers~\cite{peebles2023scalable, gupta2024photorealistic}, enabling large-scale training and improved temporal coherence. Representative examples include closed-source systems such as Sora~\cite{openai2024sora} and MovieGen~\cite{polyak2024movie}, and open-source counterparts such as CogVideoX~\cite{yang2024cogvideox}, LTXVideo~\cite {HaCohen2024LTXVideo}, Hunyuan~\cite{kong2024hunyuanvideo}, and Wan~\cite{wan2025wan}. While these bidirectional diffusion models achieve exceptional visual fidelity and realistic motion, their joint denoising process implicitly accesses future frames, violating causal ordering and precluding real-time streaming generation.

\noindent\textbf{Autoregressive Video Generation.}
To introduce temporal causality absent in bidirectional diffusion, earlier autoregressive approaches~\cite{bruce2024genie, kondratyuk2023videopoet, wang2024loong, weissenborn2019scaling, yan2021videogpt} model video synthesis as sequential prediction, producing spatiotemporal tokens conditioned on previously generated content. More recently, a complementary line combines autoregressive generation with diffusion by distilling a pretrained bidirectional teacher into a fast, causal autoregressive student. CausVid~\cite{yin2025slow} first established this paradigm via ODE initialization and a DMD~\cite{yin2024one, yin2024improved} pipeline that converts the teacher into a few-step causal student. Self-Forcing~\cite{huang2025self} addresses exposure bias by explicitly unrolling autoregressive generation during training. Subsequent works~\cite{liu2025rolling, yang2025longlive} further address error accumulation in long video generation, and LongLive~\cite{yang2025longlive} introduces a KV re-cache mechanism for prompt switching during streaming inference. Despite these advances, existing methods remain limited to text- and image-to-video generation, lacking fine-grained interactive control over scene dynamics.

\noindent\textbf{Controllable Video Generation.} Enabling generation controllability is essential for learning world models that can respond to external interventions and evolve under user actions. To this end, a wide range of work has explored various types of control representations, including structure control~\cite{xing2024tooncrafter, yang2025layeranimate, jiang2025vidsketch, pang2024dreamdance, xing2025motioncanvas}, camera control~\cite{gao2024cat3d, zheng2024cami2v, he2024cameractrl, bai2025recammaster, wu2025cat4d, yu2025trajectorycrafter, bahmani2024vd3d, yang2024direct, zheng2025vidcraft3, yu2024wonderworld}, and subject control~\cite{huang2025videomage, liu2025animateanywhere, fei2025skyreels, liu2025phantom}. Recently, Force-Prompting~\cite{gillman2025force} introduced physical force as conditioning signals, enabling control over object motion through applied forces. Although these approaches significantly enhance the semantic and geometric controllability of video diffusion models, most operate within the bidirectional diffusion framework, which limits causal consistency and hinders real-time interaction. Recent interactive systems, such as Hunyuan-GameCraft~\cite{li2025hunyuan} and Yan~\cite{ye2025yan} allow user-driven camera control during autoregressive rollout. MotionStream~\cite{shin2025motionstream} further extends this direction by introducing an efficient motion-conditioned autoregressive framework that supports trajectory control through track-based conditioning. 
While force-based control superficially resembles trajectory-based control, there are several fundamental differences, which are elaborated in Sec.~\ref{sec:diff_motionstream}.
These works mostly remain at the viewpoint or motion level, focusing on user-drawn camera movements or trajectories rather than the underlying physical dynamics that govern object behavior. In contrast, our work targets force-conditioned autoregressive generation, introducing physically grounded controllability that directly governs object interactions and scene evolution.

\noindent\textbf{Simulation-in-the-loop Video Generation.}
Another line of research~\cite{zhangPhysDreamerPhysicsBasedInteraction2024,liu2024physics3d,Cao_2024_NeuMA,jiang2025phystwin,xia2025drawerdigitalreconstructionarticulation,xia2024video2game,hsu2024autovfx,xie2024vid2sim,decoupledGaussian,lin2025omniphysgs,qiu-2024-featuresplatting,huang2024dreamphysics,xie2023physgaussian,liu2025physflow,liu2024physgen,chen2025physgen3d,tan2024physmotion,xie2025physanimator} 
uses simulators to produce physically consistent motion and leverages video diffusion models mainly for visual synthesis or refinement.
WonderPlay~\cite{li2025wonderplay} tightens this coupling by iterating between a physics solver and a video generator, using the generated video to update the simulated scene state to form a hybrid generative simulator. PhysCtrl~\cite{wang2025physctrl} learns a generative dynamics prior from large-scale simulator data and predicts 3D point trajectories as control signals for downstream video generation, reducing the need to directly run a simulator at inference time. While these simulation-connected approaches improve physical fidelity via explicit solvers or simulation-trained dynamic priors, they typically introduce additional simulation components and/or intermediate 3D trajectory representations. Our work instead focuses on force-conditioned causal autoregressive video generation directly, enabling interactive rollout without explicit simulators.

\section{Method}
\label{sec:method}
\subsection{Overview and Motivation}
\begin{figure*}[!t]
    \centering
    \includegraphics[width=0.85\linewidth]{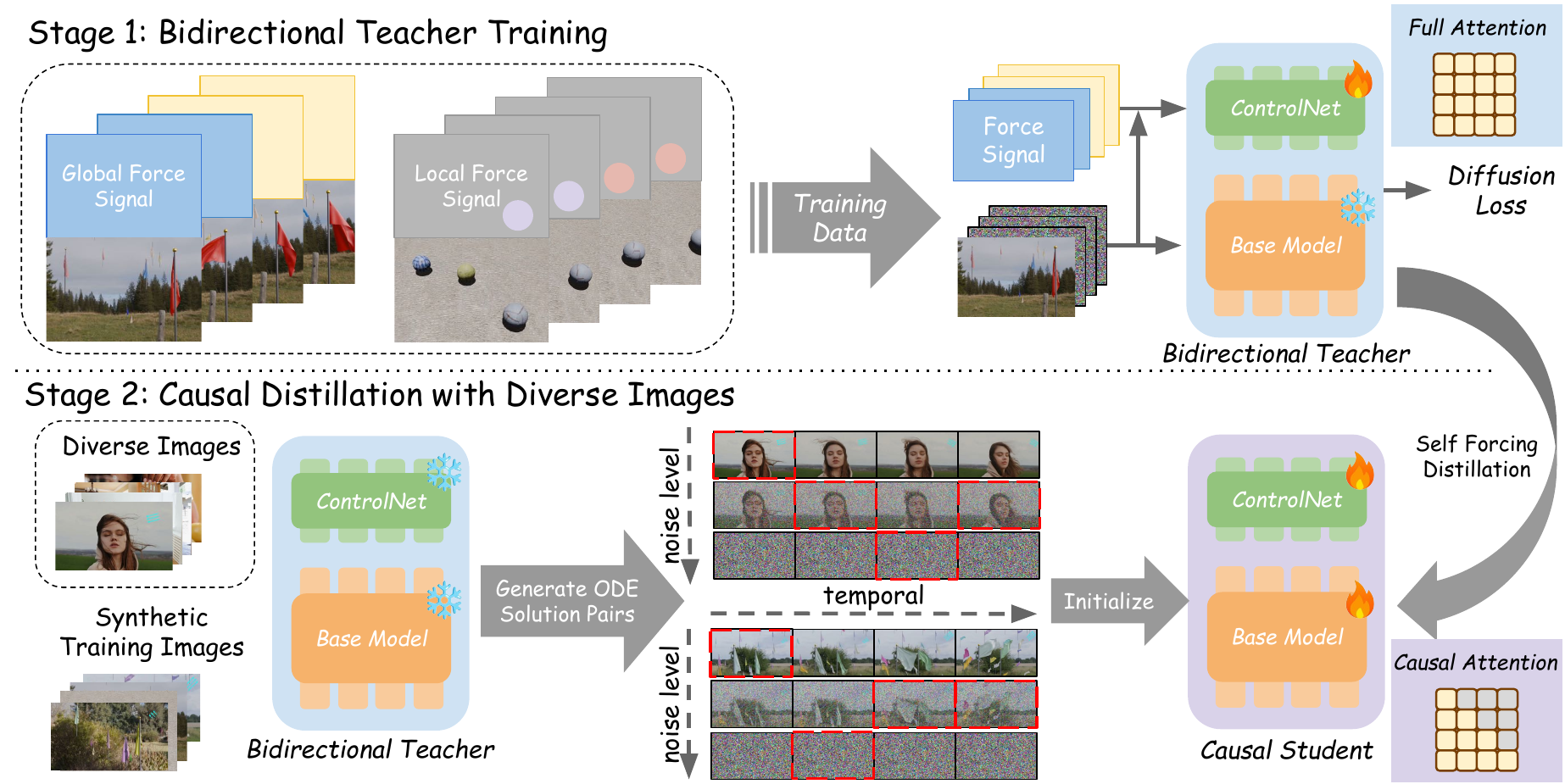}
    \Description{StreamForce two-stage training pipeline diagram: a bidirectional teacher with a unified force ControlNet, distilled into a causal student via ODE initialization plus Self-Forcing DMD distillation.}
    \caption{
        {\bf \ours} model architecture and training pipeline. \textit{Stage 1}: a bidirectional teacher is trained on synthetic force-conditioned videos using a unified force representation that supports both global and local force control with temporally changing forces (shown as color variations). \textit{Stage 2}: the teacher generates ODE solution pairs from diverse and synthetic image–force inputs. These solution pairs are then used to initialize a causal student, where different noise levels are applied to the latent chunks during initialization (illustrated by the red dashed boxes), followed by Self-Forcing DMD distillation. The resulting model supports autoregressive video generation with dynamically changing force control. 
    }
    \label{fig:pipeline}
\end{figure*}

To jointly satisfy \textbf{force interaction} and \textbf{streaming generation}, we adopt a control-before-causality paradigm that first establishes strong force-conditioned motion dynamics in a high-capacity bidirectional teacher model, and then distills this controllable behavior into a causal streaming student generator.
As illustrated in \textbf{\cref{fig:pipeline}}, \ours~consists of two stages: bidirectional teacher training with unified force representation (\cref{sec:method-1}), and causal distillation with diverse image-force trajectories (\cref{sec:method-2}).

\subsection{Bidirectional Teacher Training}
\label{sec:method-1}
\noindent\textbf{Architecture Design.}
The goal of this stage is to enable a pretrained bidirectional video generation model to respond coherently to external force inputs. 
Rather than learning motion dynamics from scratch, we leverage the strong spatiotemporal priors already embedded in large-scale video pretraining and align them with explicit force-conditioning signals.
To preserve the pretrained model’s general motion understanding, we follow Force-Prompting~\cite{gillman2025force} to adopt a separate conditioning branch~\cite{zhang2023adding} that injects force information, instead of modifying the base model parameters directly. 
This design allows the pretrained model to preserve its general motion understanding, while the control branch steers generation according to external force inputs.

\noindent\textbf{Unified Force Representation Design.}
Given this architecture, we next consider how to encode external forces for the conditioning branch.
Prior work~\cite{gillman2025force} adopts separate models and disjoint representations for global and local forces.
In contrast, we propose a unified representation that jointly models both.

A physical force at time $t$ is parameterized by its magnitude $F_t\in[0, 1]$ and direction $\theta_t\in[0, 2\pi)$. To distinguish global and local forces, we define a spatial mask $\mathbf{M}_t(u, v)\in\{0, 1\}$ over spatial coordinates $(u, v)$:
{\small
\begin{equation}
    \mathbf{M}_t(u,v) =
        \begin{cases}
            1, & \text{global force}, \\
            \mathbb I[(u-x)^2+(v-y)^2\le r^2], & \text{local force}.
        \end{cases}
\end{equation}
}

\noindent where $(x, y)$ is the initial location of the local force, and $r$ is its radius of influence. The full force tensor is then defined as
{\small
\begin{equation}
    \mathbf{f}_t(u,v) =
    \left(
    -1 + 2\mathbf{M}_t(u,v),
    -1 + 2F_t,
    \cos\theta_t,
    \sin\theta_t
    \right), 
\end{equation}
}

\noindent and stacked over time as $\mathbf{f}\in \mathbb{R}^{T\times4\times H\times W}$.
As illustrated in \textbf{\cref{fig:pipeline}}, this representation encodes both forces in a shared pixel-aligned masked force map.

\noindent\textbf{Force-Conditioned Dataset with Changing Force.} 
Following the insights of Force-Prompting~\cite{gillman2025force}, we construct a synthetic training dataset rendered in Blender in which objects of different shapes and materials are subjected to physically simulated forces such as wind fields and point interactions. These synthetic clips provide physically faithful supervision, establishing explicit correspondence between external force signals and the resulting object motion under controlled settings. 
Different from Force-Prompting~\cite{gillman2025force}, which trains separate models for each force type, we combine global and local force examples within a unified training set. 
Moreover, we introduce \textit{force-changing} settings in which the applied force varies over time in magnitude and/or direction within a single sequence. 
This exposes the model to temporally changing control signals, allowing it to learn consistent motion responses under dynamic force transitions. 
Together, this mixed training paradigm enables the bidirectional teacher to interpret spatial masks, directional and magnitude components, and time-varying forces within a single framework, forming a unified controllable model capable of handling diverse physical interventions.
We show the dataset examples in the supplementary.

\subsection{Causal Distillation with Diverse Image-Force Data}
\label{sec:method-2}
Although the bidirectional teacher learns force-motion correspondence, its joint denoising formulation relies on full-sequence conditioning and therefore cannot support causal, real-time streaming generation. 
To enable sequential synthesis where future frames depend only on previously generated content, we distill the controllable bidirectional teacher into a causal autoregressive student. 

\noindent\textbf{ODE Initialization.}
Directly training a causal student model using the DMD~\cite{yin2024one, yin2024improved} loss is often unstable due to the architectural gap between the bidirectional teacher and the causal autoregressive student. 
To mitigate this issue, we follow CausVid~\cite{yin2025slow} and adapt an ODE-based initialization strategy to stabilize training. 
Specifically, we use the pretrained bidirectional teacher to simulate reverse diffusion trajectories via an ODE solver. 
Starting from Gaussian noise samples $\{x_T^i\}_{i=1}^N$, where $N$ denotes the number of latent frames, the teacher integrates the ODE solver to produce a set of latent trajectories $\{x_t^i\}_{i=1}^N$ over denoising timesteps $t \in [T, 0]$. 
We then sample several intermediate timesteps from the timestep set used in the few-step causal generator and distillation process. The timestep is sampled independently for each latent chunk, allowing different chunks to start from different noise levels during initialization (illustrated by the red dashed boxes in \textbf{\cref{fig:pipeline}}), and train the student with a simple regression objective:
{\small
\begin{equation}
    \mathcal{L}_{\text{init}} = \mathbb{E}_{x, t_i}\|G_\phi(\{x_{t_i}^i\}_{i=1}^N, \{t_i\}_{i=1}^N, \mathbf{f}^i) - \{x_0^i\}_{i=1}^N\|_2^2,
\end{equation}
}

\noindent where $G_\phi$ denotes the causal generator initialized from the teacher’s parameters, and $\{x_0^i\}_{i=1}^N$ are the teacher's denoised latents.

Unlike prior works that rely solely on text-conditioned data for initialization, we generate two complementary types of ODE trajectories: \ding{182} \textbf{synthetic force-conditioned} and \ding{183} \textbf{diverse image-force} trajectories. 
Synthetic force-conditioned trajectories can preserve controllability learned from the teacher. 
However, relying on them alone may limit visual diversity and bias the student toward synthetic scenes. To address this, we introduce diverse image-force trajectories, which expose the model to broader visual content. Specifically, we collect diverse images from Pexels and annotate each image with the corresponding force.
This dual-source initialization transfers controllable dynamics while preventing the student from overfitting to the limited synthetic scenarios, providing a stable and well-balanced starting point for subsequent Self-Forcing~\cite{huang2025self} distillation.

\noindent\textbf{Self-Forcing Distillation.}
To enable streaming generation with external force conditioning, we distill our finetuned bidirectional teacher into a causal few-step autoregressive student model following the Self-Forcing~\cite{huang2025self} paradigm, which performs temporal autoregressive roll-out with distribution matching distillation (DMD~\cite{yin2024one, yin2024improved}). Following the chunk-wise strategy, we divide a video latent $x_t$ into $L$ chunks as $\{x_t^i\}_{i=1}^L$, where $t$ is the denoising timestep. The sampling process of the $i$-th chunk can attend to its own noisy tokens $x_t^i$ and previously clean key and value tokens stored in the KV cache $\text{KV}^i$: $\mathcal{C}^i=\{x_t^i\}\ \bigcup\ \{\text{KV}^i\}$. Formally, the denoising process is defined as: 

{\small
\begin{equation}
    x_{t_{j-1}}^i=\Psi(G_\phi(x_{t_j}^i, t_j, \mathcal{C}^i, \mathbf{f}^i), \epsilon, t_{j-1}),
\end{equation}
}

\noindent where $\Psi$ denotes the forward diffusion process, and $\epsilon$ is random Gaussian noise. After generating all $L$ chunks through self-rollout, we obtain the video latent $\hat{x}_0=\{x_0^i\}_{i=1}^L$, and apply the DMD objective to this sequence, minimizing the reverse KL divergence between our student's output distribution and the data distribution: $\mathcal{L}_\text{DMD}=\mathbb{E}_t(D_\text{KL}(p_t^{gen}||p_t^{data}))$. The gradient of the reverse KL can be approximated as:
{\small
\begin{equation}
    \nabla_\phi \mathcal{L}_\text{DMD}\approx-\mathbb{E}_{t, \hat{x}_0}\left[\left(s_\text{real}(\Psi(\hat{x}_0, t), t)-s_\text{fake}(\Psi(\hat{x}_0, t), t)\right)\cdot\frac{\partial\hat{x}_0}{\partial\phi}\right],
\end{equation}
}

\noindent where $s_\text{real}$ is the real-data score function given by the frozen teacher, and $s_\text{fake}$ is the critic trained on the generator's outputs with the standard denoising loss:
{\small
\begin{equation}
    \mathcal{L}_\text{critic}=\mathbb{E}_{t, \hat{x}_0, \epsilon}\|\epsilon_\theta(x_t, t) - \epsilon\|_2^2,
\end{equation}
}

\noindent where $\theta$ is the critic's parameter and $\epsilon_\theta$ is the predicted noise. Following \cite{yin2025slow, huang2025self}, we update the critic more frequently than the generator with a 5:1 ratio for better distribution approximation. As in ODE initialization, we perform distillation jointly on \textbf{synthetic force-conditioned data} and \textbf{diverse image-force data}, allowing the student to inherit controllable physical behavior while retaining open-domain visual generalization.

\section{Experiments}
\label{experiments}

\subsection{Implementation Details}
Our framework is built upon the text-image to video (TI2V) variants of Wan2.2 (5B), which serves as the backbone base model for bidirectional teacher training and causal distillation. Unless otherwise specified, all experiments are conducted using this base model.
We train the ControlNet branch using the rectified flow matching objective~\cite{liu2022flow, lipman2022flow}. 

For bidirectional teacher training, we generate 30K synthetic force-conditioned videos using the publicly available Force-Prompting~\cite{gillman2025force} codebase and simulation pipeline. In addition, we render another 30K synthetic videos in Blender that include force-changing scenarios with temporally varying force inputs.
For causal distillation, we further collect a diverse image dataset by downloading images from Pexels and annotating approximately 90K image-force pairs. These annotations specify force magnitude, direction, and (for local forces) spatial location, enabling supervision under broader visual contexts beyond the synthetic domain.

Additional training hyperparameters, data processing, and implementation specifics are provided in the supplementary.

\subsection{Evaluation Setup}
\noindent\textbf{Baselines.}
We compare \ours~against three representative baselines.
\ding{182} \textbf{Wan2.2 5B TI2V}~\cite{wan2025wan}, used directly as a text-to-video inference baseline by providing the initial image together with rule-based force-related textual prompts.
\ding{183} \textbf{Force-Prompting}~\cite{gillman2025force}, which we retrain on the Wan2.2 5B backbone for fair comparison, as the original implementation is built upon CogVideoX~\cite{hong2022cogvideo}.
\ding{184} \textbf{Kling 1.5 with Motion Brush}, 
which supports localized trajectory-based interaction, and is evaluated primarily on local force scenarios.

\noindent\textbf{Perceptual Evaluation Protocols.} 
For perceptual evaluation, we construct an image--force test set consisting of 40 cases using images downloaded from Pexels. The cases cover four scenarios: local force preservation, local force change, global force preservation, and global force change. For each case, all compared methods receive the same initial image and force condition to generate video outputs.
We conduct a human user study in which participants evaluate the generated videos according to three criteria: \ding{182} \textbf{force adherence} (how well the motion aligns with the specified force), \ding{183} \textbf{physical plausibility} (whether the motion follows realistic physical behavior), and \ding{184} \textbf{visual quality}. 

\noindent\textbf{Physics-IQ Evaluation Protocols.}
To quantitatively evaluate physical consistency, we follow the Physics-IQ benchmark~\cite{motamed2025physics} and compute its metrics on a real-world test set collected by ourselves. Specifically, we record 40 video cases under controlled force conditions, which serve as ground truth for comparison.
Since the Physics-IQ evaluation requires precise alignment between generated motion and real-world trajectories, we restrict the setting to \textit{force preservation} cases where the applied force direction remains fixed throughout the sequence. 
For each case, we provide all compared methods with the same initial frame and corresponding force input. To ensure fair comparison, all evaluated models are configured to produce outputs with identical resolution and frame-rate settings required for Physics-IQ metric computation.

\subsection{Main Results}

\begin{table*}[!t]
\centering
\setlength{\tabcolsep}{6pt}
\caption{\textbf{Perceptual Evaluation}. \textit{FT}: Force Type. We report the preserving force in the top block and the changing force in the bottom block. The best and second-best results are marked in \textbf{bold} and \uline{underline}, respectively.}
\vspace{-0.8em}
\label{tab:perceptual}
\resizebox{0.85\textwidth}{!}{
\begin{tabular}{l l c ccc ccc}
\toprule
\multicolumn{3}{c}{} &
\multicolumn{3}{c}{\textbf{Global Force}} &
\multicolumn{3}{c}{\textbf{Local Force}} \\
\cmidrule(lr){4-6} \cmidrule(lr){7-9}

\textbf{FT} &
\textbf{Method} &
\textbf{Streaming} &
\textbf{Force Adh.} &
\textbf{Real. Phys.} &
\textbf{Vis. Qual.} &
\textbf{Force Adh.} &
\textbf{Real. Phys.} &
\textbf{Vis. Qual.} \\
\midrule

\multirow{4}{*}{\rotatebox[origin=c]{90}{Preserve}}
& ForcePrompt & \xmark & \textbf{74.2\%} & \uline{47.3\%} & 38.8\% & \uline{48.8\%} & 32.3\% & 30.8\% \\
& Kling Brush & \xmark & - & - & - & 44.2\% & \uline{45.8\%} & \uline{52.3\%} \\
& Wan 2.2 TI2V & \xmark & 20.8\% & 31.2\% & \uline{40.4\%} & 15.0\% & 17.7\% & 17.7\% \\
& \textbf{Ours} & \cmark & \uline{64.2\%} & \textbf{65.8\%} & \textbf{68.5\%} & \textbf{81.9\%} & \textbf{71.9\%} & \textbf{69.2\%} \\
\midrule

\multirow{4}{*}{\rotatebox[origin=c]{90}{Change}}
& ForcePrompt & \xmark & \uline{32.7\%} & \uline{38.5\%} & 33.1\% & 6.2\% & 25.0\% & 19.2\% \\
& KlingBrush & \xmark & - & - & - & \uline{34.2\%} & \uline{35.8\%} & \uline{43.1\%} \\
& Wan 2.2 TI2V & \xmark & 7.3\% & 28.1\% & \uline{35.4\%} & 4.2\% & 12.3\% & 13.1\% \\
& \textbf{Ours} & \cmark & \textbf{86.5\%} & \textbf{77.3\%} & \textbf{76.9\%} & \textbf{80.4\%} & \textbf{64.6\%} & \textbf{62.7\%} \\
\bottomrule
\end{tabular}
}
\end{table*}

\noindent\textbf{Perceptual Evaluation.}
We evaluate perceptual controllability and video realism under both \textit{force preservation} and \textit{force change} settings. The quantitative results are summarized in \textbf{\cref{tab:perceptual}}.

Across both global and local force scenarios, \ours~predominantly achieves the highest scores in all three criteria under both preservation and changing force settings. The improvements are particularly pronounced in the force-changing cases (bottom half of \textbf{\cref{tab:perceptual}}), where the models must respond to changed force control signals. Notably, ~\ours~also surpasses Kling 1.5 Motion Brush in several local force scenarios, despite being a fully streaming autoregressive model. The results suggest that our design effectively transfers fine-grained controllability into a causal generation framework without sacrificing perceptual realism. 

\begin{figure*}[!t]
    \centering
    \includegraphics[width=1.0\linewidth]{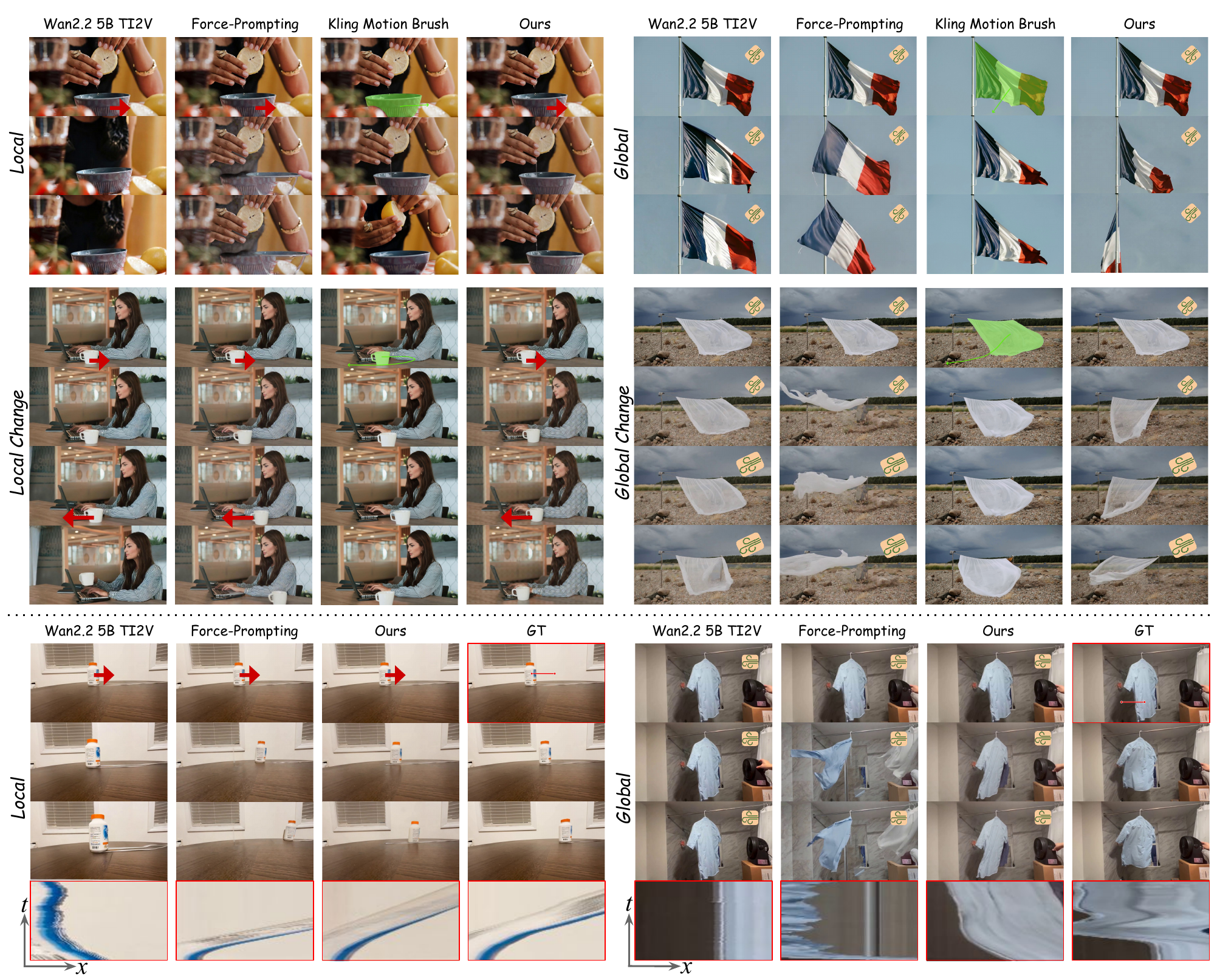}
    \Description{Qualitative video comparison grid: baselines versus StreamForce under force-conditioned scenarios, with x-t slices visualizing motion.}
    \caption{
        {\bf Visual Comparison}. Baseline methods often exhibit weak or mismatched motion responses, \ours~ produces motion dynamics that more faithfully follow the intended force input with stable visual appearance and plausible physicality. We visualize x-t (space-time) slices on our recorded force-conditioned video in the last row to demonstrate the motion.
        \textit{Zoom in for details.}
    }
    \label{fig:qualitative}
\end{figure*}

Qualitative comparisons are shown in \textbf{\cref{fig:qualitative}} (top). While baselines often exhibit weak or mismatched motion, \ours~produces dynamics that more faithfully follow the input force with stable appearance and plausible physicality.

\begin{table*}[!t]
\centering
\setlength{\tabcolsep}{6pt}
\caption{\textbf{Physics-IQ Benchmark}~\cite{motamed2025physics}. We record 40 video cases under controlled force conditions, which serve as ground truth for comparison. \textit{S.T. IoU}: SpatialTemporal IoU. \textit{W.S. IoU}: Weighted Spatial IoU. Arrows ($\uparrow$ or $\downarrow$) indicate whether a higher or lower value is preferred. The best results are marked in \textbf{bold}.}
\label{tab:physiq}
\vspace{-0.8em}
\resizebox{0.85\textwidth}{!}{
\begin{tabular}{l ccccc ccccc}
\toprule
\multicolumn{1}{c}{} &
\multicolumn{5}{c}{\textbf{Global Force}} &
\multicolumn{5}{c}{\textbf{Local Force}} \\
\cmidrule(lr){2-6} \cmidrule(lr){7-11}

\multirow{2}{*}{\textbf{Method}} &
\textbf{Spatial} &
\textbf{S.T.} & \textbf{W.S.} & \textbf{MSE} & \textbf{Total} &
\textbf{Spatial} &
\textbf{S.T.} & \textbf{W.S.} & \textbf{MSE} & \textbf{Total} \\

& \textbf{IoU $\uparrow$} &
\textbf{IoU $\uparrow$} & \textbf{IoU $\uparrow$} & \textbf{$\downarrow$} & \textbf{Score $\uparrow$} &
\textbf{IoU $\uparrow$} &
\textbf{IoU $\uparrow$} & \textbf{IoU $\uparrow$} & \textbf{$\downarrow$} & \textbf{Score $\uparrow$} \\

\midrule
ForcePrompt & \textbf{0.118} & 0.496 & \textbf{0.033} & 0.012 & 40.87 & \textbf{0.217} & 0.373 & \textbf{0.188} & 0.006 & 44.29 \\
Wan 2.2 TI2V & 0.061 & 0.450 & 0.021 & 0.015 & 37.92 & 0.077 & 0.300 & 0.048 & 0.017 & 35.21\\
\midrule
\textbf{Ours} & 0.091 & \textbf{0.532} & 0.028 & \textbf{0.011} & \textbf{40.99} & 0.168 & \textbf{0.542} & 0.147 & \textbf{0.005} & \textbf{46.31} \\
\bottomrule
\end{tabular}
}
\end{table*}

\noindent\textbf{Physics-IQ Evaluation.}
Quantitative Physics-IQ results are summarized in~\textbf{\cref{tab:physiq}}. Across both global and local force scenarios, \ours~achieves the highest aggregated Physics-IQ score and leads on the spatiotemporal IoU and motion-error (MSE) metrics. While Force-Prompting attains higher pixel-level Spatial IoU and Weighted Spatial IoU, \ours~produces motion that more closely matches the ground-truth dynamics over time, reflecting more physically consistent behavior.

Qualitative comparisons are shown in \textbf{\cref{fig:qualitative}} (bottom). In both local and global cases, baseline methods exhibit noticeable deviations from the ground truth trajectories, whereas \ours~produces motion patterns that more closely follow the recorded physical behavior. The accompanying x-t visualizations further highlight that our model generates spatiotemporal motion structures more consistent with ground truth, reflecting more accurate force-driven dynamics.
\subsection{Ablation Studies}
\begin{figure*}[!t]
\centering
\Description{Two side-by-side ablation tables: Physics-IQ scores comparing force representations and training data variants, and a perceptual-study table comparing force-changing supervision.}

\begin{minipage}[t]{0.55\textwidth}
  \centering
  \setlength{\tabcolsep}{6pt}
  \captionsetup{type=table}
  \caption{\textbf{Ablations with PhysicsIQ Score}. \textit{F-Prompt Rep.}: use Force Prompting~\cite{gillman2025force} representation. \textit{Separate}: train separate models for global and local force. \textit{w/o Diverse}: remove diverse data during distillation. Arrows ($\uparrow$ or $\downarrow$) indicate whether a higher or lower value is preferred.}
  \label{tab:physiq_unified_rep}
  \resizebox{1.0\textwidth}{!}{
    \begin{tabular}{l c c|l c c}
    \toprule
    \multicolumn{1}{c}{\textbf{Method}} &
    \multicolumn{1}{c}{\textbf{Global}$\uparrow$} &
    \multicolumn{1}{c}{\textbf{Local}$\uparrow$} &
    \multicolumn{1}{c}{\textbf{Method}} &
    \multicolumn{1}{c}{\textbf{Global}$\uparrow$} &
    \multicolumn{1}{c}{\textbf{Local}$\uparrow$}\\
    \midrule
    
    F-Prompt Rep. & 40.87 & 44.29 &
    w/o Diverse & \multirow{2}{*}{27.82} & \multirow{2}{*}{20.50} \\
    
    Separate & 41.79 & 47.38 &
    Data & & \\
    
    \midrule
    \textbf{Ours} (teacher) & 42.13 & 49.99 &
    \textbf{Ours} & 40.99 & 46.31 \\
    \bottomrule
    \end{tabular}
  }
\end{minipage}
\hspace{0.02\textwidth}
\begin{minipage}[t]{0.26\textwidth}
  \centering
  \setlength{\tabcolsep}{6pt}
  \captionsetup{type=table}
  \caption{\textbf{Ablations with Perceptual Study}. \textit{w/o Change Data}: remove force-changing data during teacher model training.}
  \label{tab:changing_force}
  \resizebox{1.0\textwidth}{!}{
    \begin{tabular}{l c c}
      \toprule
      \multicolumn{1}{c}{\textbf{Method}} &
      \multicolumn{1}{c}{\textbf{Global}} &
      \multicolumn{1}{c}{\textbf{Local}} \\
      \midrule
      w/o Change & \multirow{2}{*}{15.2\%} & \multirow{2}{*}{0\%} \\
      Data & & \\
      \midrule
      \textbf{Ours} & 71.7\% & 88.6\%\\
      \bottomrule
    \end{tabular}
  }
\end{minipage}

\end{figure*}

We conduct ablation studies on force representation, unified training, diverse-data distillation, and force-changing supervision.

\noindent\textbf{Different Force Representations.}
We first compare our unified force representation with the Force-Prompting formulation (denoted as F-Prompt Rep.), which encodes global and local forces using separate designs. As shown in \textbf{\cref{tab:physiq_unified_rep}} (left block), our representation achieves improved performance on both global and local force scenarios. These results indicate that our pixel-aligned masked force map representation can effectively model both global and local force conditions within a single framework. Importantly, this unified design enables shared training across force types, laying the foundation for the joint training strategy.

\begin{figure*}[!t]
    \centering
    \includegraphics[width=1.0\linewidth]{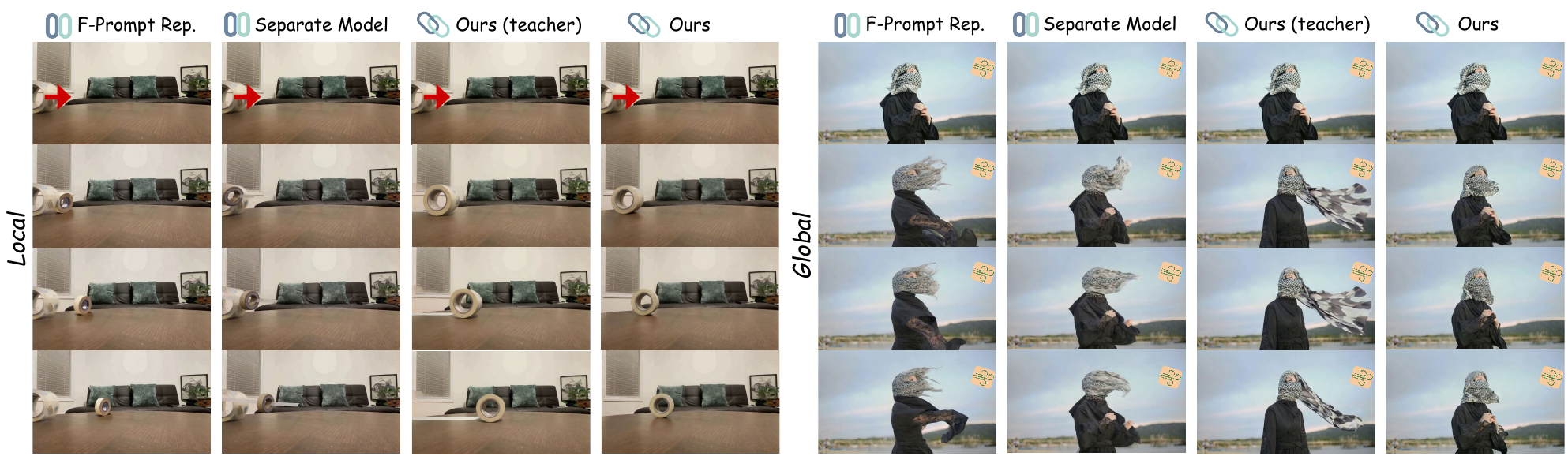}
    \Description{Ablation study qualitative comparison contrasting the Force-Prompting representation, separate-models baseline, and our unified-representation teacher and student.}
    \caption{
        {\bf Ablation Studies}. \textit{F-Prompt Rep.}: use Force-Prompting~\cite{gillman2025force} representation. \textit{Separate}: train separate models for global and local force. 
        Using a unified force representation with joint training (Ours-teacher) maintains generation quality while enabling cross-force generalization. This capability also carries over to the autoregressive model (Ours).
        \includegraphics[height=1em]{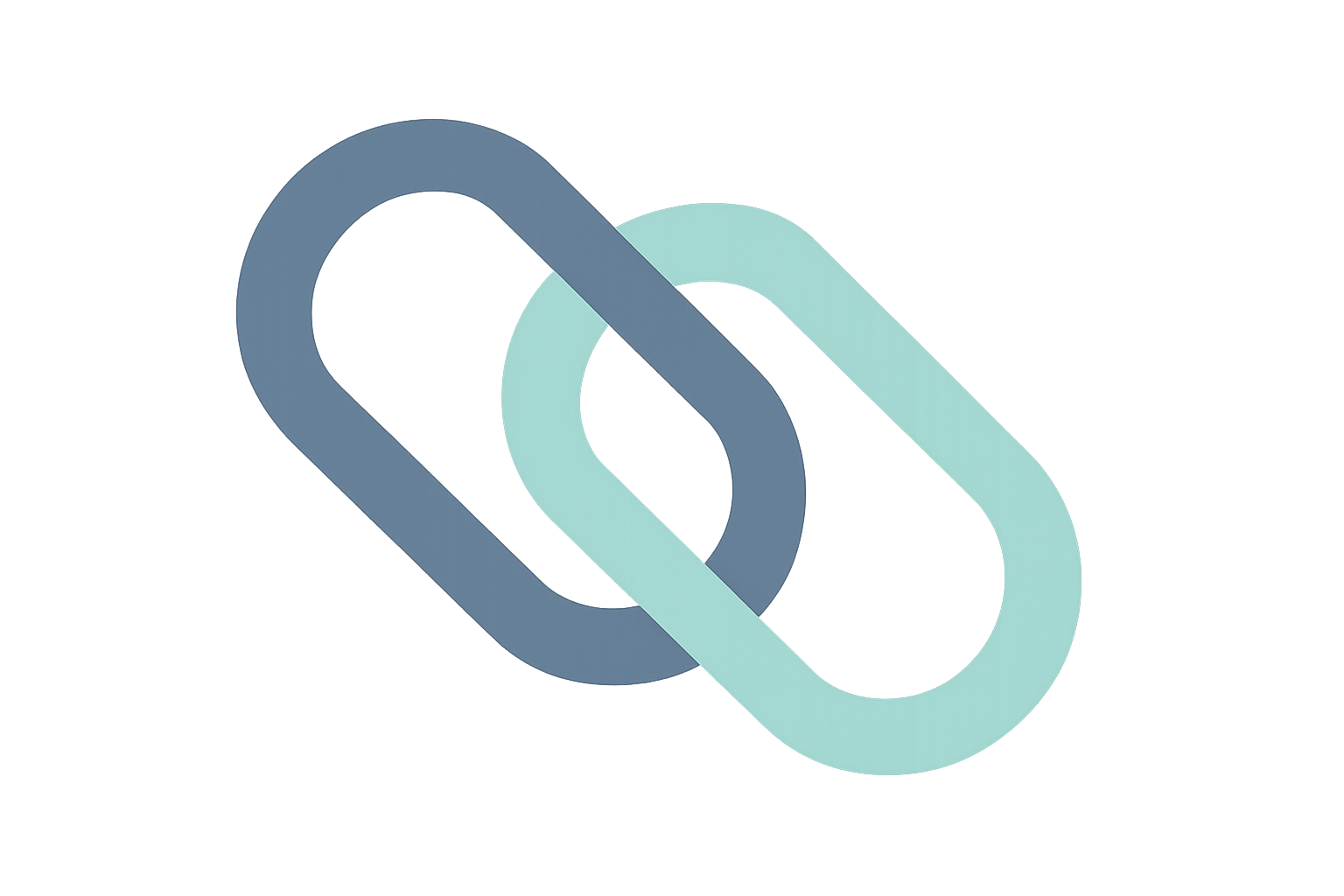} indicates unified model for local and global force while \includegraphics[height=1em]{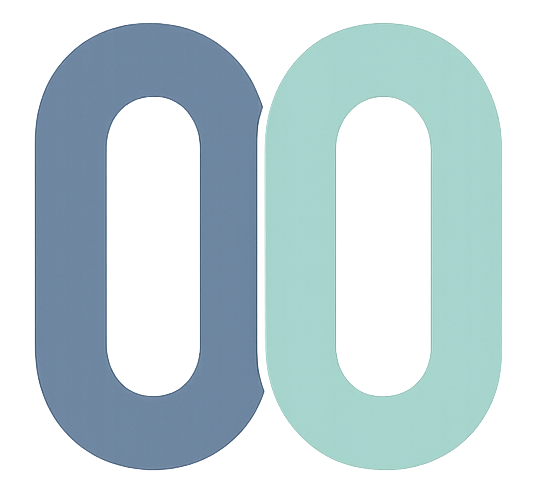} indicates a separate model.
    }
    \label{fig:ablations_1}
\end{figure*}

\noindent\textbf{Unified vs. Separate Training.}
Building upon the unified representation, we next investigate whether training separate models for global and local forces is preferable to joint training within a single model. As shown in \textbf{\cref{tab:physiq_unified_rep}} (left block), unified training achieves stronger overall performance across both force types compared to training separate models. This suggests that a shared representation encourages cross-force generalization and leads to more consistent motion modeling by promoting shared motion dynamic priors between global and local settings. Qualitative comparisons are shown in \textbf{\cref{fig:ablations_1}}. The unified model produces more coherent and spatially aligned motion patterns than both the Force-Prompting representation and separate models.

\noindent\textbf{Diverse Image-Force Data in Distillation.}
We next analyze the role of diverse image-force data during causal distillation. As shown in \textbf{\cref{tab:physiq_unified_rep}} (right block), removing diverse data leads to clear performance degradation across both global and local force scenarios. When distilled using only synthetic data, the model tends to overfit to limited training distributions and struggles to generalize to broader visual content. Incorporating diverse image–force pairs during distillation alleviates this issue and consistently improves overall performance, highlighting the importance of diverse visual supervision for streaming controllable generation.

\begin{figure*}[!t]
    \centering
    \includegraphics[width=1.0\linewidth]{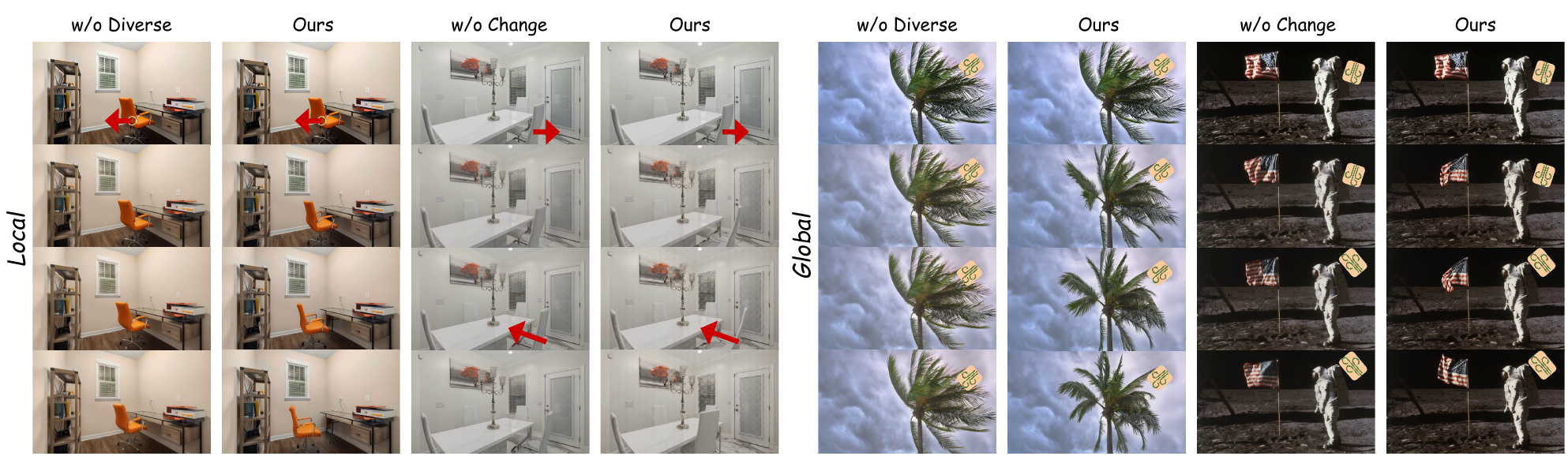}
    \Description{Ablation study qualitative comparison showing the effect of removing diverse data during distillation and removing force-changing training data.}
    \caption{
        {\bf Ablation Studies}. Removing diverse data during distillation (\textit{w/o Diverse}) leads to fewer motion patterns and reduced adaptability to complex scenes. 
        Removing force-changing data (\textit{w/o Change}) causes the model to largely ignore force updates and produce little response after the force changes.
    }
    \label{fig:ablations_2}
\end{figure*}

Interestingly, we observe that the bidirectional teacher trained solely on synthetic data maintains reasonable generalization, whereas the causal student distilled without diverse supervision exhibits a more pronounced performance drop. This suggests that autoregressive distillation imposes additional modeling constraints, making diverse supervision particularly important for maintaining generalization in the streaming setting. Qualitative comparisons between our full model and the variant without diverse data distillation are shown in \textbf{\cref{fig:ablations_2}}. Without diverse data, the model exhibits fewer motion patterns and reduced adaptability to complex visual content.

\noindent\textbf{Force-Conditioned Dataset with Changing Force.}
Finally, we evaluate the necessity of force-changing supervision during training. As shown in \textbf{\cref{tab:changing_force}}, removing force-changing data leads to a dramatic performance drop under change-force scenarios. Although force inputs can still be modified at inference time, models trained only on force-preservation cases fail to respond consistently to dynamic force updates.
As illustrated in \textbf{\cref{fig:ablations_2}}, removing force-changing supervision causes the model to largely ignore force updates, exhibiting little to no response after the force is changed.

\section{Discussion}
\begin{figure*}[!t]
\centering
\begin{minipage}[t]{0.49\textwidth}
    \centering
    \includegraphics[width=\linewidth]{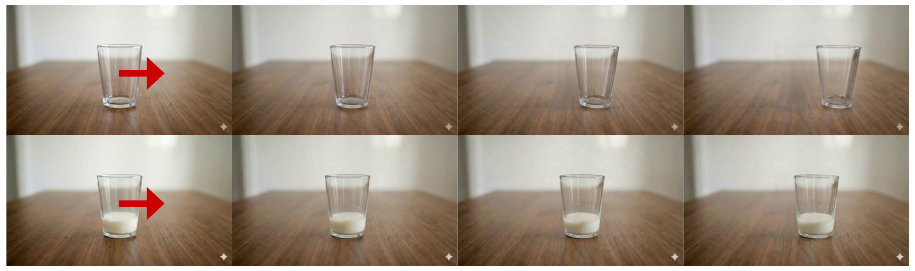}
    \Description{Two side-by-side video sequences in which the same horizontal force is applied to two glasses, one empty and one containing milk; the milk-filled glass moves more slowly.}
    \captionof{figure}{\textbf{Mass-aware motion behavior.} Under the same horizontal force, the glass containing milk moves more slowly than the empty glass, reflecting the expected relationship between object mass and acceleration. \textit{Zoom in for details.}}
    \label{fig:mass}
\end{minipage}
\hfill
\begin{minipage}[t]{0.49\textwidth}
    \centering
    \includegraphics[width=\linewidth]{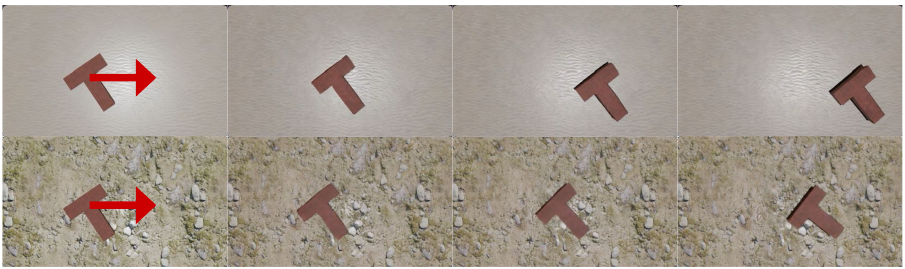}
    \Description{Two side-by-side video sequences in which the same horizontal force is applied to the same T-shaped object on two surfaces with different friction; the object travels a shorter distance on the higher-friction surface.}
    \captionof{figure}{\textbf{Friction-aware motion behavior.} Under the same horizontal force applied to the same T-shaped object, the object travels a shorter distance on a higher-friction surface than on a smoother one, reflecting how friction opposes motion and dissipates energy. \textit{Zoom in for details.}}
    \label{fig:friction}
\end{minipage}
\end{figure*}

\noindent\textbf{Emergent Intuitive Physics.} We observe behaviors broadly consistent with \emph{intuitive physics}~\cite{battaglia2013simulation, wu2015galileo, wu2016physics101}: qualitative reasoning about everyday physical interactions. As shown in \textbf{\cref{fig:mass}}, under the same horizontal force a milk-filled glass tends to move more slowly than an empty one, in line with heavier objects accelerating less. As shown in \textbf{\cref{fig:friction}}, the same force applied to the same T-shape produces a shorter displacement on a rougher-looking surface than on a smoother one. These observations suggest that \ours~captures coarse visual cues rather than a quantitatively faithful physics model.

\begin{figure*}[!t]
    \centering
    \includegraphics[width=\linewidth]{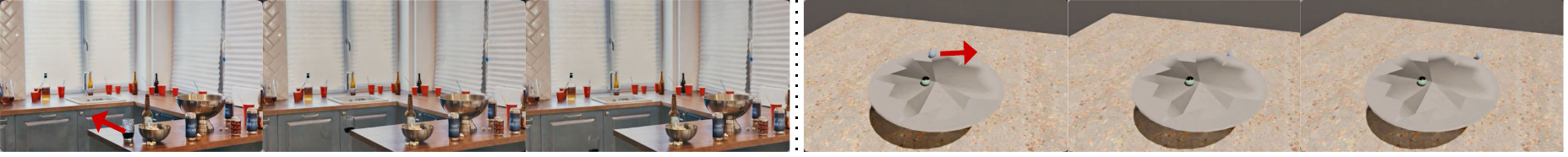}
    \Description{Video sequence in which a force pushes an object across a tabletop until it falls off the edge and lands on the ground.}
    \vspace{-3mm}
    \caption{\textbf{Object Falling.} A force pushes the object across a table; once it passes the edge, it falls under gravity. \textit{Zoom in for details.}}
    \label{fig:falling_bouncing}
\end{figure*}

\noindent\textbf{Object Falling.}
\ours~also captures basic gravitational dynamics. As illustrated in \textbf{\cref{fig:falling_bouncing}}, an applied force pushes an object across a table toward its edge; once it passes the edge, the generated motion follows a downward trajectory consistent with gravity. This behavior is not explicitly conditioned on, but rather emerges from the spatiotemporal priors inherited from the pretrained video model, suggesting that \ours~retains a coarse intuitive physics beyond direct force-driven interaction.

\begin{figure*}[!t]
    \centering
    \includegraphics[width=\linewidth]{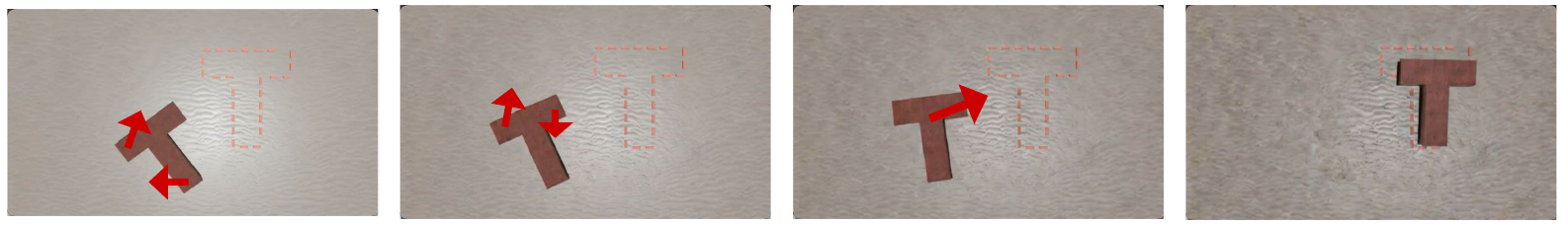}
    \Description{T-pushing manipulation example showing two simultaneous local forces applied to different parts of a T-shaped object, producing coordinated translation and rotation.}
    \vspace{-3mm}
    \caption{\textbf{T-Pushing Manipulation.} Applying two local forces simultaneously to different parts of a T-shaped object produces coordinated translation and rotation that drive the object toward a target position.}
    \label{fig:t_pushing}
\end{figure*}

\noindent\textbf{Multi-Force and Part-Level Interaction.}
Our unified force representation naturally supports multiple forces applied at different parts of the same object. As a special case, we consider T-pushing, a canonical robotics manipulation task in which a T-shaped object must be pushed to a target position. 
As shown in \textbf{\cref{fig:t_pushing}}, \ours~handles this scenario by applying two local forces simultaneously to different parts of the T-shape, producing coherent motion that drives the object toward the target position. 
This demonstrates that our streaming force control supports multi-force, part-level manipulation, opening potential applications in interactive robotics scenarios.

\label{sec:diff_motionstream}
\noindent\textbf{Differences from trajectory-control based works~\cite{shin2025motionstream,geng2024motionprompting}.}
 While force-based control superficially resembles trajectory-based control, there are several fundamental differences.
First, global forces such as fluid dynamics are naturally described as forces, yet are difficult or impossible to represent using trajectories alone. 
Second, the effect of an applied force inherently depends on an object’s physical properties, such as mass or material composition, a dependency that is absent when specifying motion or location (e.g., the same force produces a larger displacement in lighter objects, as shown in \cref{fig:mass}).

\section{Conclusion}
\label{conclusion}
In this work, we introduce \ours~, a streaming video generation framework that enables interaction through continuous force inputs. By combining a unified force representation with a force-aware distillation pipeline, our approach bridges force-controllable video generation and causal autoregressive synthesis within a single model that supports both global and local, time-varying forces. Extensive experiments demonstrate that \ours~ achieves state-of-the-art performance in force-conditioned video generation, moving generative video models closer to interactive world models.

\FloatBarrier
{\small
\bibliographystyle{ieeenat_fullname}
\bibliography{main}
}

\FloatBarrier
\clearpage
\section*{\Large{Appendix}}
\setcounter{section}{0}
\setcounter{figure}{0}
\setcounter{table}{0}
\makeatletter 
\renewcommand{\thesection}{\Alph{section}}
\renewcommand{\theHsection}{\Alph{section}}
\renewcommand{\thefigure}{A\arabic{figure}} %
\renewcommand{\theHfigure}{A\arabic{figure}} %
\renewcommand{\thetable}{A\arabic{table}}
\renewcommand{\theHtable}{A\arabic{table}}
\makeatother

\renewcommand{\thetable}{A\arabic{table}}
\setcounter{equation}{0}
\renewcommand{\theequation}{A\arabic{equation}}

\appendix

\section{Perceptual Study Details}
We conduct three perceptual user studies to evaluate different aspects of \ours: \ding{182} overall comparisons with baselines, \ding{183} sensitivity to force magnitude, and \ding{184} responsiveness to changing forces. All studies are conducted through an online survey interface where participants compare videos generated under indentical input conditions. Participants may select one or multiple video results that best satisfy the evaluation criterion or choose \textit{None} if none of the results are satisfactory. To reduce potential bias, the order of the compared methods is randomly shuffled for each question and for each participant.

\subsection{Overall Comparisons with Baselines}
\begin{figure*}[t]
\centering
\begin{subfigure}{0.48\linewidth}
    \centering
    \includegraphics[width=\linewidth]{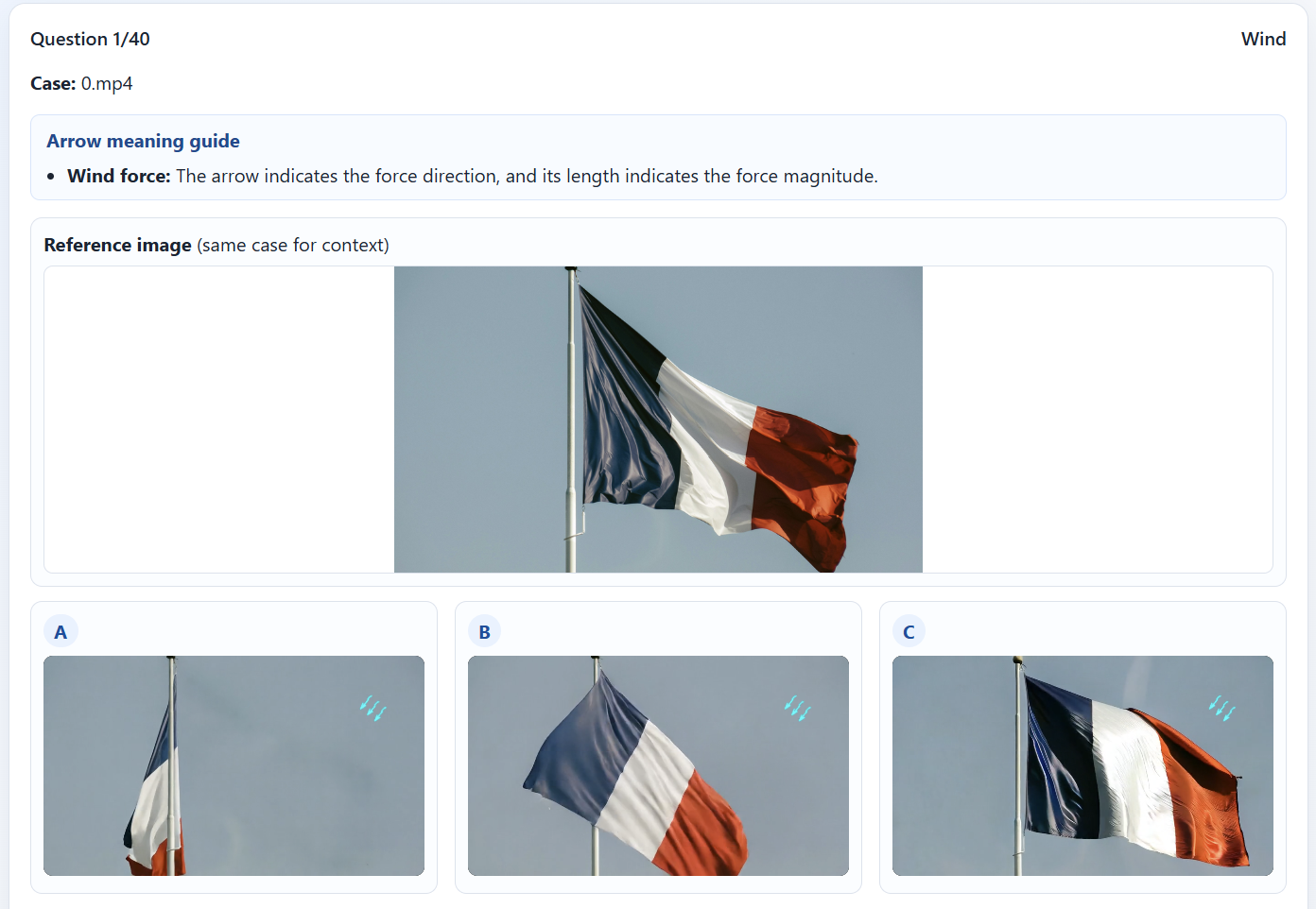}
    \Description{Perceptual study video comparison panel screenshot.}
    \caption{Video comparison panel}
\end{subfigure}
\hfill
\begin{subfigure}{0.48\linewidth}
    \centering
    \includegraphics[width=\linewidth]{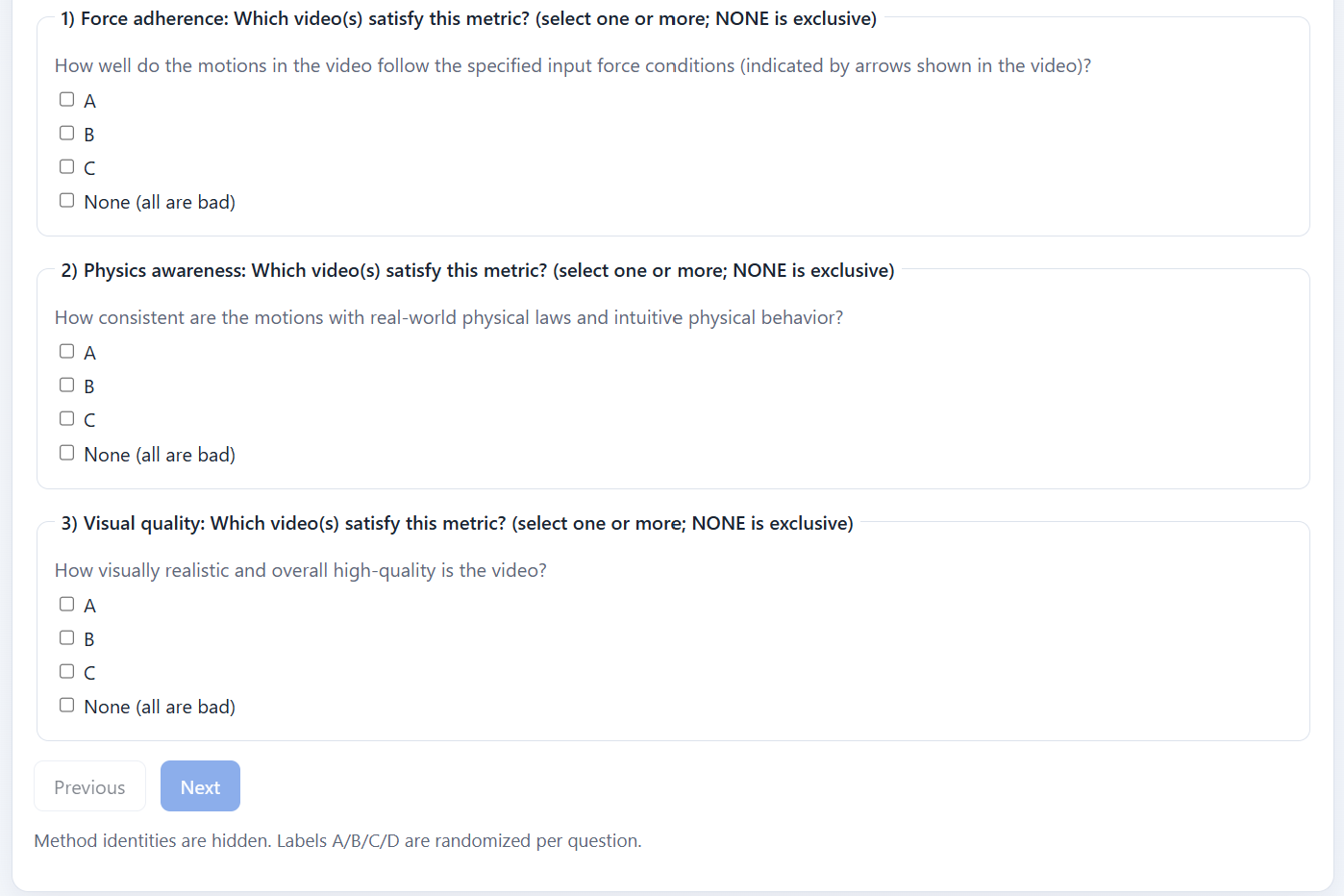}
    \Description{Perceptual study evaluation questions screenshot.}
    \caption{Evaluation questions}
\end{subfigure}

\caption{\textbf{Perceptual study interface for overall comparisons with baselines.}
Left: generated videos from different methods shown to participants for comparison.
Right: evaluation questions for force adherence, physics awareness, and visual quality.
Arrows are overlaid on the videos to indicate the input force direction and magnitude.}
\label{fig:perceptual_main}
\end{figure*}

This study evaluates the overall controllability and perceptual quality of \ours~generated videos compared with baseline methods. We construct a test set consisting of four types of force control scenarios: \textit{global force}, \textit{local force}, \textit{changing global force}, and \textit{changing local force}. Each category contains 10 evaluation cases, resulting in 40 cases in total. A total of 26 participants completed the study. The aggregated results are reported in Tab. 1 in the main paper.
Participants evaluate the generated videos according to three criteria: \textbf{force adherence}, \textbf{physics awareness}, and \textbf{visual quality}. Force adherence measures how well the motion follows the specified force direction and location; physics awareness evaluates whether the motion appears consistent with intuitive real-world physical behavior; and visual quality reflects the overall perceptual realism of the generated video.
An example interface of the perceptual study is shown in \textbf{\cref{fig:perceptual_main}}. For each evaluation case, participants are first shown the initial image used for video generation as a reference. Videos generated by different methods are then presented for comparison. To clearly communicate the input force conditions, arrows are overlaid on the generated videos, where the arrow direction indicates the force direction and the arrow length represents the force magnitude. Participants select one or more videos that satisfy each evaluation criterion or choose \textit{None} if none of the results are satisfactory.

\subsection{Sensitivity to Force Magnitude}
\begin{figure*}[t]
    \centering
    \includegraphics[width=\linewidth]{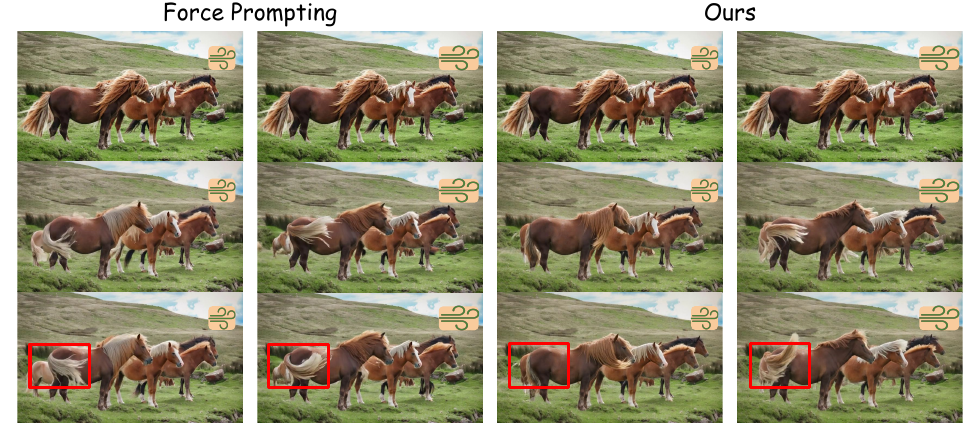}
    \Description{Magnitude response comparison: two horse videos generated under weaker versus stronger wind forces, showing different motion amplitudes in mane and tail.}
    \caption{\textbf{Magnitude Response Comparisons}. Ours responds more clearly to force magnitude differences than the baseline. \textit{Zoom in for details.}}
    \label{fig:magnitude}
\end{figure*}

\begin{figure*}[t]
    \centering
    \includegraphics[width=0.8\linewidth]{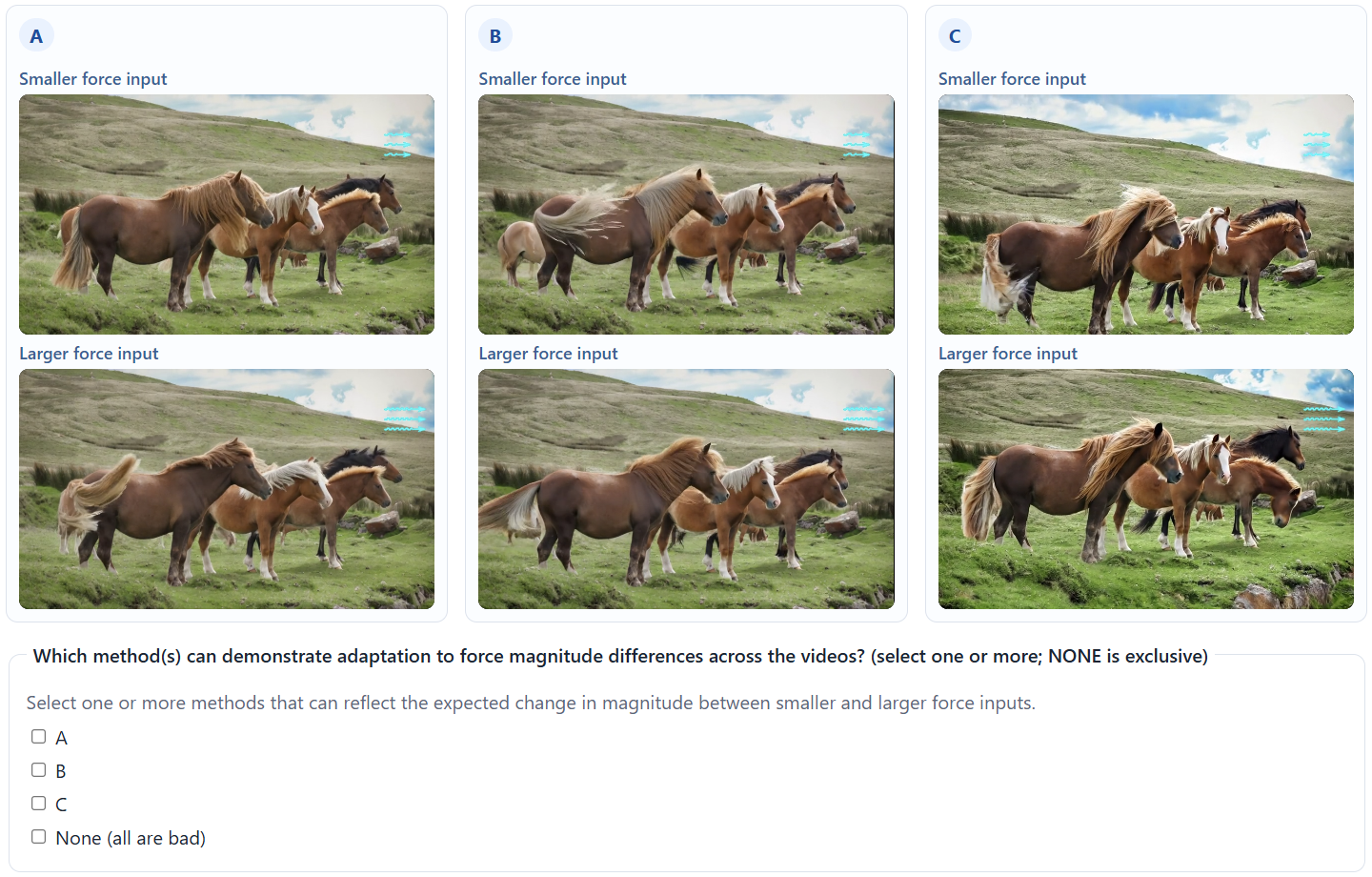}
    \Description{Perceptual study interface screenshot for the force-magnitude sensitivity task.}
    \caption{\textbf{Perceptual study interface for force magnitude sensitivity.} Each method produces two videos with different force magnitudes applied to the same initial image. The top row corresponds to a smaller force input, and the bottom row corresponds to a larger force input. }
    \label{fig:perceptual_magnitude}
\end{figure*}

While motion direction can be directly evaluated from the trajectory of moving objects, assessing whether motion strength reflects the magnitude of the applied force is more challenging. To specifically examine this capability, we conduct an additional perceptual study focusing on the model's sensitivity to \textbf{force magnitude differences}.
In this study, participants compare videos generated under two force magnitudes applied to the same initial image: a \textit{smaller force input} and a \textit{larger force input}. For each method, two videos corresponding to these two force magnitudes are presented. Participants are asked to determine which method(s) can clearly demonstrate adaptation to the magnitude difference, i.e., whether the generated motion becomes stronger when the force magnitude increases.
An example interface of this study is shown in \textbf{\cref{fig:perceptual_magnitude}}. Each method is presented as a column containing two videos: the top row corresponds to the smaller force input, and the bottom row corresponds to the larger force input. Arrows overlaid on the videos indicate the input force direction and magnitude. Participants select the method(s) whose generated motions best reflect the expected difference between smaller and larger forces, or choose \textit{None} if none of the methods exhibit clear magnitude adaptation.
As summarized in \textbf{\cref{tab:magnitude}}, \ours~substantially outperforms baseline methods in reflecting the intended magnitude variations.
A representative example is shown in \textbf{\cref{fig:magnitude}}: when a horse is subjected to weaker versus stronger wind forces, \ours~produces visibly different motion amplitudes in the mane and tail (highlighted by the red boxes), whereas Force-Prompting generates nearly identical motion under both conditions. This demonstrates that our model responds more consistently to changes in force magnitude.
\begin{table}[t]
\centering
\setlength{\tabcolsep}{6pt}
\caption{\textbf{Magnitude Response Perceptual Study.} Ours more accurately reflects the intended motion differences between smaller and larger forces compared to the baselines.}
\label{tab:magnitude}
\begin{tabular}{l c c}
  \toprule
  \multicolumn{1}{c}{\textbf{Method}} &
  \multicolumn{1}{c}{\textbf{Global}} &
  \multicolumn{1}{c}{\textbf{Local}} \\
  \midrule
  Wan2.2 TI2V & 0\% & 0\% \\
  ForcePrompt & 16.7\% & 20.0\% \\
  \midrule
  \textbf{Ours} & \textbf{88.9\%} & \textbf{78.0\%} \\
  \bottomrule
\end{tabular}
\end{table}

\subsection{Responsiveness to Changing Forces}
\begin{figure*}[t]
\centering
\begin{subfigure}{0.48\linewidth}
    \centering
    \includegraphics[width=\linewidth]{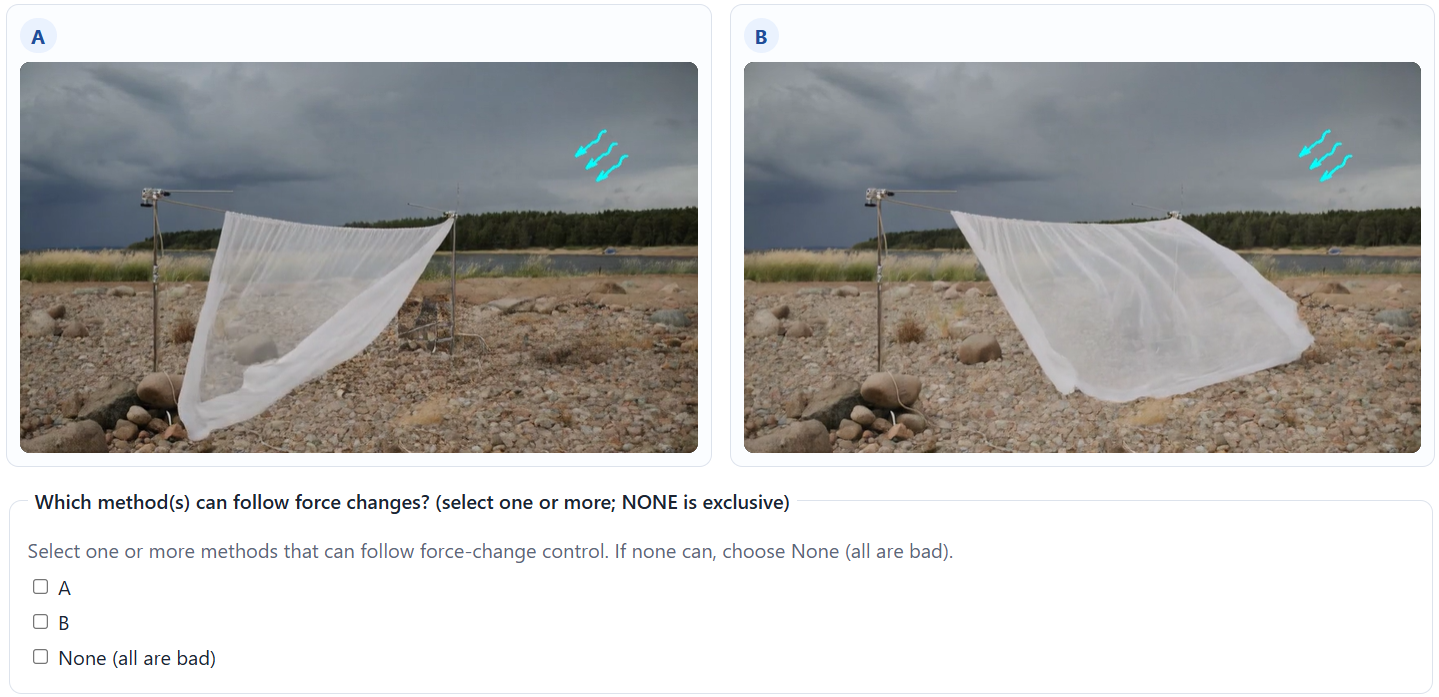}
    \Description{Perceptual study interface screenshot showing video frames before the force is changed.}
    \caption{Before change}
\end{subfigure}
\hfill
\begin{subfigure}{0.48\linewidth}
    \centering
    \includegraphics[width=\linewidth]{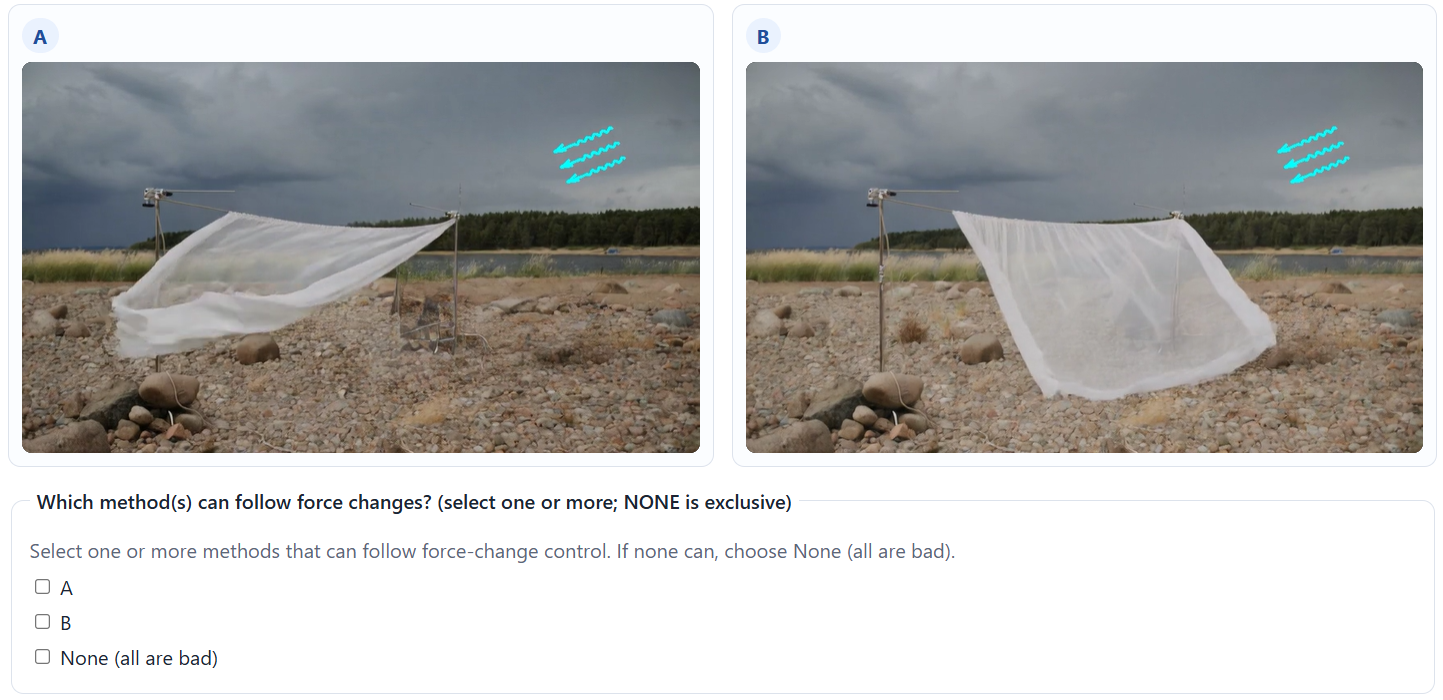}
    \Description{Perceptual study interface screenshot showing video frames after the force is changed.}
    \caption{After change}
\end{subfigure}

\caption{\textbf{Perceptual study interface for force change responsiveness.}
Two variants each generate a video under a force-changing condition. To illustrate the task setup, we show frames from both videos at two different time steps: \textit{before change} and \textit{after change}. The force change can be inferred from the updated arrow.}
\label{fig:perceptual_change}
\end{figure*}

We further evaluate whether the model can correctly respond to changes in the applied force during video generation. In this study, we compare two variants of \ours: the final model trained with force-changing supervision and an ablated version trained without force-changing data, where force changes are only applied during inference. An example of the interface is shown in \textbf{\cref{fig:perceptual_change}}. Two videos generated by the compared variants are displayed for each case. The applied force changes during the sequence, which is indicated by the updated arrow. To illustrate this setup, we show frames from both videos at two time steps corresponding to \textit{before change} and \textit{after change}. Participants are asked to determine which method better follows the updated force input after the change occurs, or select \textit{None} if neither method responds appropriately. The aggregated results of this study are reported in Tab. 4 in the main paper.

\section{Physics-IQ Benchmark Details}

To evaluate physical consistency, we adopt the metrics from the Physics-IQ benchmark~\cite{motamed2025physics}. 
The evaluation cases provided by Physics-IQ are primarily designed for text-to-video scenarios that test general physical reasoning. 
Since our task focuses on force-conditioned video generation, we instead construct our own evaluation set tailored to explicit force inputs while keeping the Physics-IQ metric computation unchanged.

\begin{figure*}[t]
    \centering
    \includegraphics[width=1.0\linewidth]{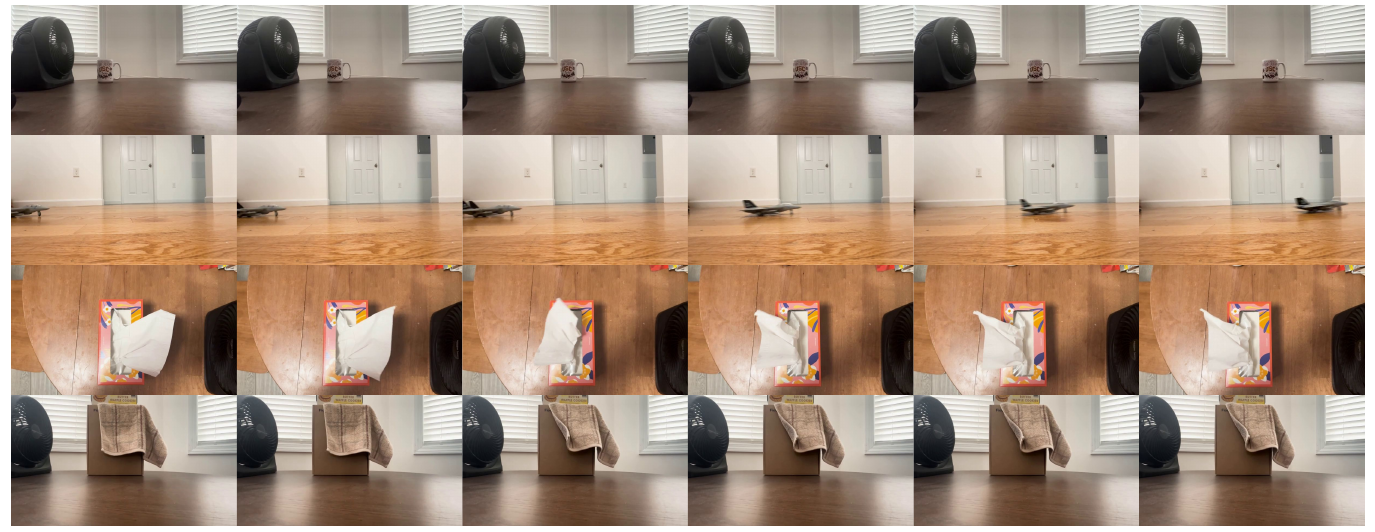}
    \Description{Recorded real-world video sequences used as ground truth for the Physics-IQ evaluation, showing global and local force cases.}
    \caption{\textbf{Examples of recorded sequences used for Physics-IQ evaluation.}}
    \label{fig:physicsiq_examples}
\end{figure*}

We record 40 real-world video sequences under controlled force conditions, covering both global- and local-force scenarios.
Global forces are generated using a fan to simulate wind, while local forces are applied by dragging objects with a line or poking movable objects to induce motion. 
Each sequence lasts 5 seconds and is recorded using an iPhone at 60 FPS. 
Following the protocol used in Physics-IQ, we uniformly subsample the recorded videos to obtain 80 frames per sequence. 
All generated videos are produced at a resolution of 832$\times$480 and evaluated over the same 80-frame duration (corresponding to 5 seconds at 16 FPS). Example sequences from our recorded evaluation set are shown in \textbf{\cref{fig:physicsiq_examples}}.

To ensure fair comparison across methods, for each initial image and force input we generate 5 videos using different random seeds and report the mean Physics-IQ score across these runs. 
Results are reported separately for global-force and local-force scenarios in Tabs. 2 and 3 in the main paper.

We follow the Physics-IQ evaluation protocol and use their released code to compute all metrics. 
The evaluation pipeline is minimally adapted to load our ground-truth and generated videos, while the metric definitions and computation procedures remain unchanged.

\begin{figure*}[t]
    \centering
    \includegraphics[width=1.0\linewidth]{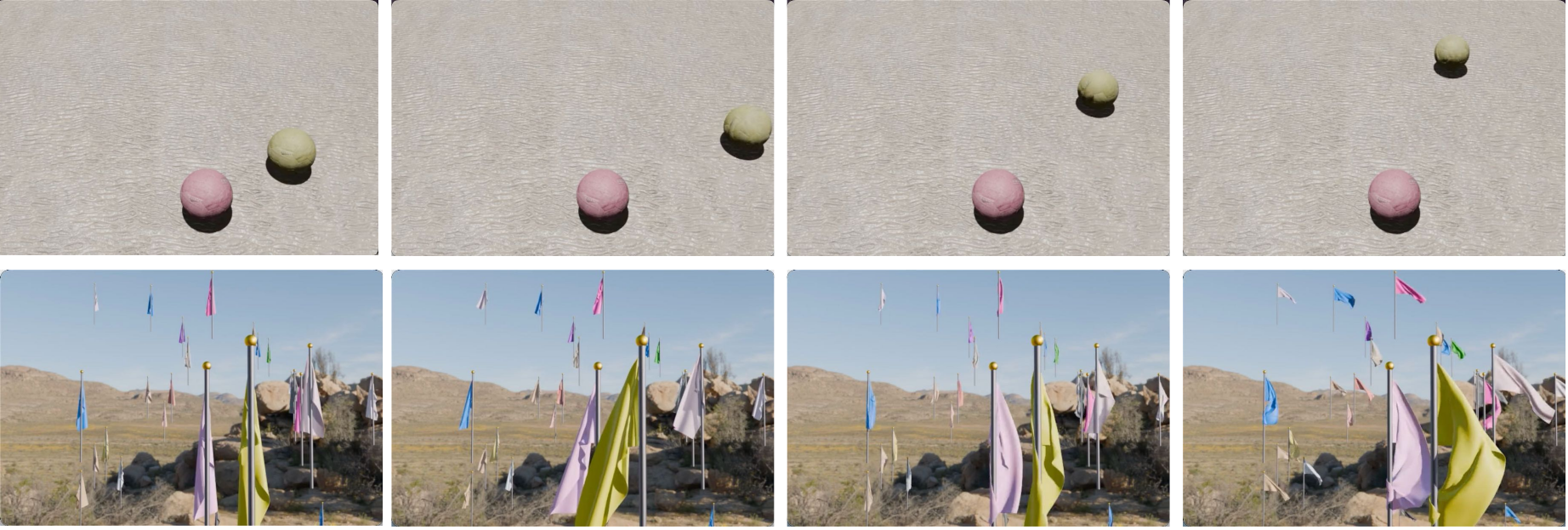}
    \Description{Examples of synthetic force-conditioned video frames used to train the bidirectional teacher model.}
    \caption{\textbf{Examples of synthetic data for bidirectional teacher training.}}
    \label{fig:train_data}
\end{figure*}

\section{Dataset Details}
\label{sec:dataset-details}
\subsection{Synthetic Data for Bidirectional Teacher Training}
To train the bidirectional teacher model, we generate synthetic force-conditioned videos using the data generation pipeline from Force-Prompting~\cite{gillman2025force}. The pipeline uses Blender to simulate object motion under controlled physical forces.
Following their setup, we generate two types of force interactions. For \textit{global forces}, wind fields are applied to cloth flags to simulate wind-driven motion. For \textit{local forces}, a ball placed on a plane is poked to produce localized force interactions that induce motion.

To support the force-changing scenarios studied in this work, we extend the Force-Prompting data generation scripts to include a \textit{force change} setting. For global forces, the wind direction and magnitude are updated at a specified time step to a new set of randomly sampled values. For local forces, the ball is poked again after the change, with a different direction and magnitude applied to the same object. This allows the generated sequences to reflect dynamic force conditions where the applied force changes during the motion.

The resulting synthetic dataset provides paired supervision consisting of the initial image, force inputs, and the corresponding motion trajectories, which are used to train the bidirectional teacher to learn force-consistent motion dynamics. 
We show example synthetic data in~\textbf{\cref{fig:train_data}}.

\begin{figure*}[t]
    \centering
    \includegraphics[width=0.8\linewidth]{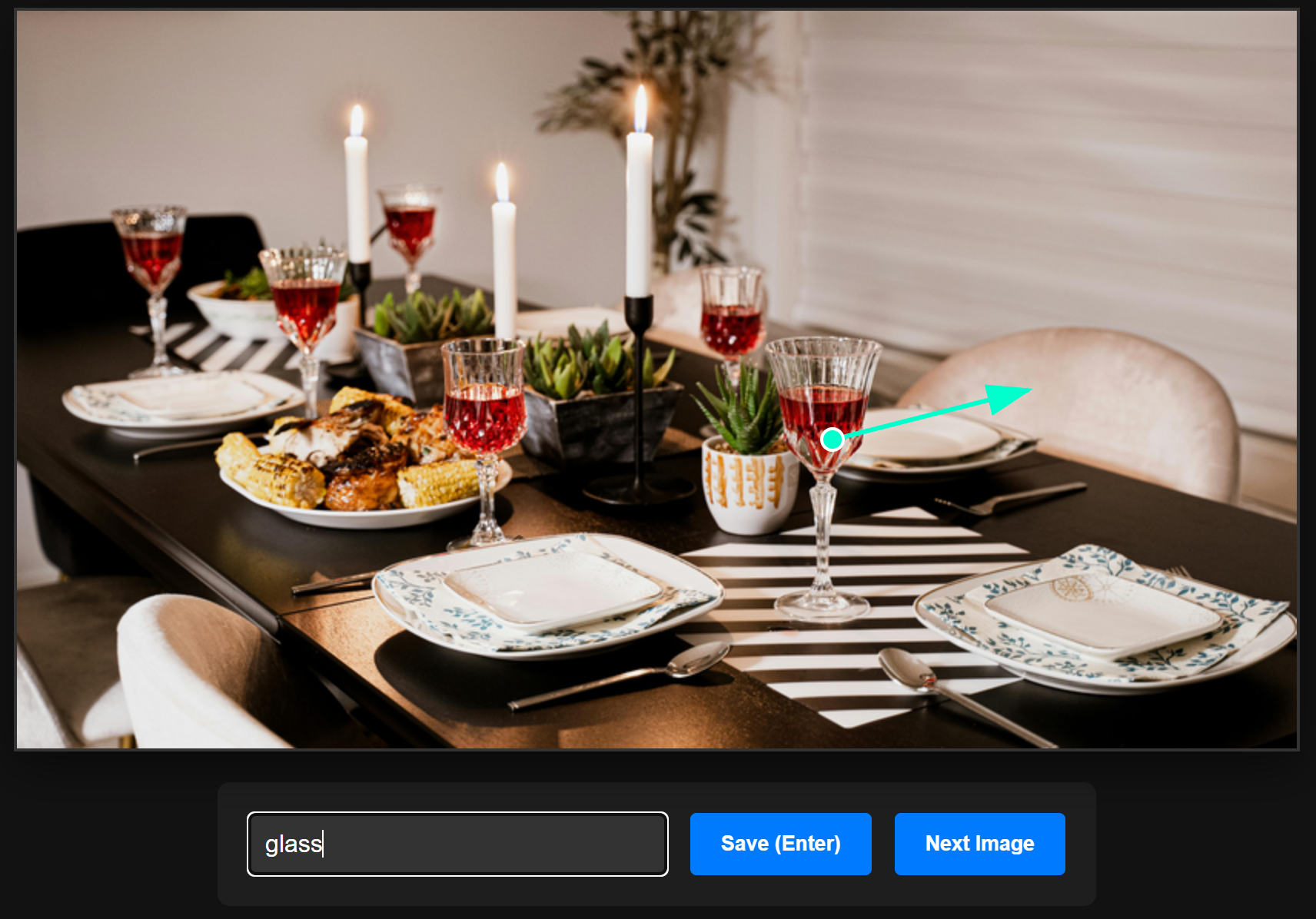}
    \Description{Screenshot of the local-force annotation interface used by annotators.}
    \caption{\textbf{Local force annotation interface.} Annotators select the target object and drag an arrow to indicate the intended motion direction. The object name, the pixel location of the click, and the arrow direction are recorded to define the local force input.}
    \label{fig:local_force_ui}
\end{figure*}

\subsection{Diverse Data for Causal Distillation}
To preserve the visual diversity of the pretrained video generator during causal distillation, we construct a diverse image–force dataset using real-world images. Relying solely on synthetic Blender data may bias the model toward limited object categories and scene layouts, while diverse images help maintain broader visual priors. 

We collect images using the Pexels API by querying a wide range of scene and object categories. For each image, we construct corresponding force inputs to form image-force pairs. For global forces, the force direction and magnitude are randomly sampled and applied to the entire scene. For local forces, we annotate movable objects using a lightweight annotation interface, as shown in \textbf{\cref{fig:local_force_ui}}. Annotators first click on the target object and then drag an arrow to indicate the intended motion direction. The selected object name, the pixel location of the click, and the arrow direction are recorded to define the local force input. For both global and local force settings, we generate a text caption using Qwen3-VL~\cite{bai2025qwen3} to ensure that the textual prompt is consistent with the image content and the intended force interaction. Importantly, the captions do not revewal the force direction or magnitude to avoid leaking supervision signals into the text input. For local force cases, the caption only includes the name of the manipulated object while omitting any information about the location.

Given the constructed image-force data pairs, we use the bidirectional teacher model to generate corresponding ODE solution pairs that serve as the training data for ODE initialization to achieve a stable starting point for Self-Forcing distillation~\cite{huang2025self}. Since randomly generated image-force combinations may occasionally produce static or implausible motion, we further employ Qwen3-VL to filter out cases where the generated videos remain largely static. The remaining samples are used to train the causal student model.

\noindent\textbf{Global force caption generation.} We prompt Qwen3-VL with the following prompt to use it to generate the captions for diverse global force samples:
\begin{tcolorbox}[enhanced, breakable, colback=gray!10, colframe=gray!40, boxrule=0.5pt]
\textbf{Objective}: \textbf{Give a highly descriptive video caption based on input image and user input}. As an expert, delve deep into the image with a discerning eye, leveraging rich creativity, meticulous thought. When describing the details of an image, include appropriate dynamic information to ensure that the video caption contains reasonable actions and plots. If user input is not empty, then the caption should be expanded according to the user's input.

\textbf{Note}: The input image is the first frame of the video, and the output video caption should describe the motion starting from the current image. Don't contain camera transitions, screen switching, or perspective shifts.

\textbf{Wind Constraint (Important)}:
The scene involves wind blowing. You MUST acknowledge the presence of wind in the description.
However, you MUST NOT specify or imply:

- wind direction (e.g., left, right, upward, sideways, from any compass direction),

- wind strength or intensity (e.g., strong, weak, gentle, violent),

- wind speed or magnitude,

- any quantitative or comparative description of the wind.

Only mention the existence of wind in a neutral, non-specific way,
such as "a breeze is present" or "the wind affects the scene",
without giving any details that could determine direction or strength.

\textbf{Answering Style}:
Answers should be comprehensive, conversational, and use complete sentences. The answer should be in English no matter what the user's input is. Provide context where necessary and maintain a certain tone. Begin directly without introductory phrases like "The image/video showcases" "The photo captures" and more.

\textbf{Output Format}: "[highly descriptive image caption here]"
\end{tcolorbox}

\noindent\textbf{Local force caption generation.}
We prompt Qwen3-VL with the following prompt to use it to generate the captions for diverse local force samples:
\begin{tcolorbox}[enhanced, breakable, colback=gray!10, colframe=gray!40, boxrule=0.5pt]
\textbf{Objective}: \textbf{Give a highly descriptive video caption based on input image and user input}. As an expert, delve deep into the image with a discerning eye, leveraging rich creativity, meticulous thought. When describing the details of an image, include appropriate dynamic information to ensure that the video caption contains reasonable actions and plots. If user input is not empty, then the caption should be expanded according to the user's input.

\textbf{Note}: The input image is the first frame of the video, and the output video caption should describe the motion starting from the current image. Don't contain camera transitions, screen switching, or perspective shifts.

**Force Constraint (Important)**:
The scene involves an external force causing an object to move. You MUST acknowledge this in the description.
However, you MUST NOT specify or imply:

- object location in the frame (e.g., left, right, center, foreground, background),

- force direction,

- force strength or intensity,

- force magnitude or speed,

- angle, coordinates, or any numeric/quantitative details.

Only describe that the object is being moved by an external push/pull in a neutral, non-quantitative way.

\textbf{Answering Style}:
Answers should be comprehensive, conversational, and use complete sentences. The answer should be in English no matter what the user's input is. Provide context where necessary and maintain a certain tone. Begin directly without introductory phrases like "The image/video showcases" "The photo captures" and more.

\textbf{Output Format}: "[highly descriptive image caption here]"
\end{tcolorbox}

\section{Training Details}
\subsection{Bidirectional Teacher Training}
We initialize our bidirectional teacher model from the pretrained Wan2.2 TI2V model~\cite{wan2025wan}. The ControlNet~\cite{zhang2023adding} branch is initialized by copying the first half of the transformer layers from the pretrained backbone. During teacher training, we optimize only the ControlNet parameters while keeping the backbone weights frozen in order to preserve the pretrained model's visual generation capabilities.
We train our teacher model on the synthetic force-conditioned dataset using 8 H200 GPUs for 10K optimization steps. We use AdamW optimizer with learning rate $1\times10^{-6}$, $\beta_1=0$, $\beta_2=0.999$, and weight decay $0.01$. Training is performed with a per-GPU batch size of 1 (global batch size of 8) using bfloat 16 mixed precision. To improve training efficiency, videos are processed at a spatial resolution of $416\times240$, which is half of the final generation resolution in both dimensions. Each training sample contains 21 latent frames, corresponding to 81 decoded video frames. 

\subsection{Causal Distillation}
\noindent\textbf{ODE initialization.} Before the final distillation training, we perform an ODE initialization stage to initialize (pretrain) a causal student model using trajectories generated by the bidirectional teacher. The student model is initialized from the bidirectional teacher weights, but is trained with causal temporal attention masks following prior causal video generation approaches~\cite{yin2025slow,huang2025self}. Unlike the teacher training stage, during ODE initialization we optimize both the backbone and ControlNet parameters so that the model can adapt to the causal attention masking used by the student architecture. Training data consists of ODE solution pairs generated by the bidirectional teacher from both the synthetic dataset and diverse image-force dataset described in \cref{sec:dataset-details}. During training, samples are drawn with a synthetic-to-diverse ratio of $1:2$. Training is performed on 16 H200 GPUs with a per-GPU batch size of 2 and graident accumulation of 4, resulting in a global batch size of 128 for 10K steps. We use the AdamW optimizer with learning rate $2\times10^{-6}$, $\beta_1=0$, $\beta_2=0.999$, and weight decay 0.01. ODE latents are processed at the target generation resolution of $832\times480$ with 21 latent frames. We used a fixed diffusion timestep index schedule of $[1000, 750, 500, 250, 0]$ to construct the ODE trajectories sued for training. 

\noindent\textbf{Self-Forcing style distillation.}
After ODE initialization, we further train the causal student model using the Self-Forcing distillation framework~\cite{huang2025self}. In this stage, the student model serves as the generator, and is jointly optimized with a critic model (fake score function, initialized using the teacher) following the DMD-style objective~\cite{yin2024one,yin2024improved}. We use the AdamW optimizer for both models, with a learning rate of $2\times10^{-6}$ for the generator and $4\times10^{-7}$ for the critic. The generator and critic are updated with a $1:5$ ratio, following prior distillation practice. 
Training is performed on 8 H200 GPUs with a per-GPU batch size of 1 at a spatial resolution of $832\times480$. We train for 3000 steps in total. We additionally maintain an exponential moving average (EMA) of the generator parameters during this stage, starting after the first 200 training steps.

\section{Long Video Generation}
To support long-horizon video generation, we adopt the inference strategy introduced in Rolling-Forcing~\cite{liu2025rolling}. During inference, instead of strictly denoising frames one-by-one, we employ a rolling diffusion window that jointly denoises multiple consecutive frames with local bidirectional attention. At the same time, we maintain a KV cache that consists of two parts: \ding{182} a temporal context cache that stores recent frames to preserve short-term consistency, and \ding{183} a global context cache that retains key-value states of initial frames as long-range anchors. This inference design effectively suppresses error accumulation and maintains global coherence during long video generation. 

\section{Limitations}
\noindent\textbf{2D only force representation.} Our current formulation represents forces within the image plane and does not explicitly model forces acting along the depth direction (i.e., toward or away from the camera). While many common scenarios such as wind or lateral pushes can be reasonably captured with this formulation, out-of-plane forces and fully three-dimensional physical interactions are not explicitly represented. 

\noindent\textbf{Limited diversity of force types.} The force interactions considered in this work focus on mechanical contact forces (e.g., pushes and pulls) and bulk wind, both of which can be described directly by Newtonian mechanics. Other physical force types, such as \emph{magnetic} forces, electrostatic forces, buoyancy, or other non-contact / field-based interactions, are not covered by our current force representation or training data. While our model can also support multi-force, part-level interactions in special cases (e.g., the T-pushing example in the discussion), this capability does not yet generalize to arbitrary objects or to arbitrary numbers of contact points; the local-force setup primarily targets single-point interactions during training. A more comprehensive physically grounded generation system should reliably reason about forces applied at different locations of an object. Expanding the range of supported force types (including non-contact forces such as magnetic and electrostatic) and multi-contact configurations would further improve the realism and generality of the framework.

\noindent\textbf{Limited object materials under local force.} The objects targeted by local forces in our current training data are predominantly rigid or articulated (e.g., a glass being pushed, a drawer being pulled). Non-rigid materials such as fluids (water splashing, smoke being stirred), elastic or deformable objects (cloth, rubber, dough), and granular media (sand, pebbles) are not explicitly covered by the local-force setup. As a result, applying a local force to such objects may yield motion that is plausible only as rigid-body translation, rather than the richer deformable or fluid response that the underlying physics would dictate. Broader coverage would require local-force training data spanning a wider range of material categories and the corresponding deformable/fluid dynamics.

\noindent\textbf{Limited modeling of object-object interactions.} Our approach primarily focuses on modeling how individual objects respond to externally applied forces. 
Complex physical interactions involving multiple objects, such as collisions, stacking, or chained motion between objects, are not explicitly modeled in the current system. 
Handling richer multi-object dynamics would likely require incorporating stronger physical priors or simulation-aware training signals.

\end{document}